\documentclass[10pt,twocolumn,letterpaper]{article}

\usepackage[pagenumbers]{cvpr}

\usepackage{cuted}

\usepackage{colortbl}
\definecolor{resultfirst}{RGB}{255,153,153}
\definecolor{resultsecond}{RGB}{255,204,153}
\definecolor{resultthird}{RGB}{255,246,170}
\newcommand{\firstscore}[1]{\cellcolor{resultfirst}#1}
\newcommand{\secondscore}[1]{\cellcolor{resultsecond}#1}
\newcommand{\thirdscore}[1]{\cellcolor{resultthird}#1}

\usepackage{microtype}

\definecolor{cvprblue}{rgb}{0.21,0.49,0.74}
\usepackage[hypertexnames=false,breaklinks,colorlinks,allcolors=cvprblue]{hyperref}

\title{TSIR: Ray-Traced Triangle Splats for Inverse Rendering}
\author{%
  Wentao Yang\textsuperscript{1,2,4} \enspace
  Chentao Shen\textsuperscript{1,2} \enspace
  Zhengqi Zhang\textsuperscript{3,4} \enspace
  Bolin Chen\textsuperscript{4} \enspace
  Yan Ye\textsuperscript{4} \enspace
  Xiangru Huang\textsuperscript{2}\\
  {\normalsize
    \textsuperscript{1}Zhejiang University \quad
    \textsuperscript{2}Westlake University \quad
    \textsuperscript{3}Tsinghua University \quad
    \textsuperscript{4}Alibaba DAMO Academy
  }
}
\hypersetup{
  pdftitle={\thetitle},
  pdfauthor={Wentao Yang, Chentao Shen, Zhengqi Zhang, Bolin Chen, Yan Ye, Xiangru Huang}
}

\begin{document}
\maketitle
\begin{strip}
    \centering
    \includegraphics[width=\linewidth]{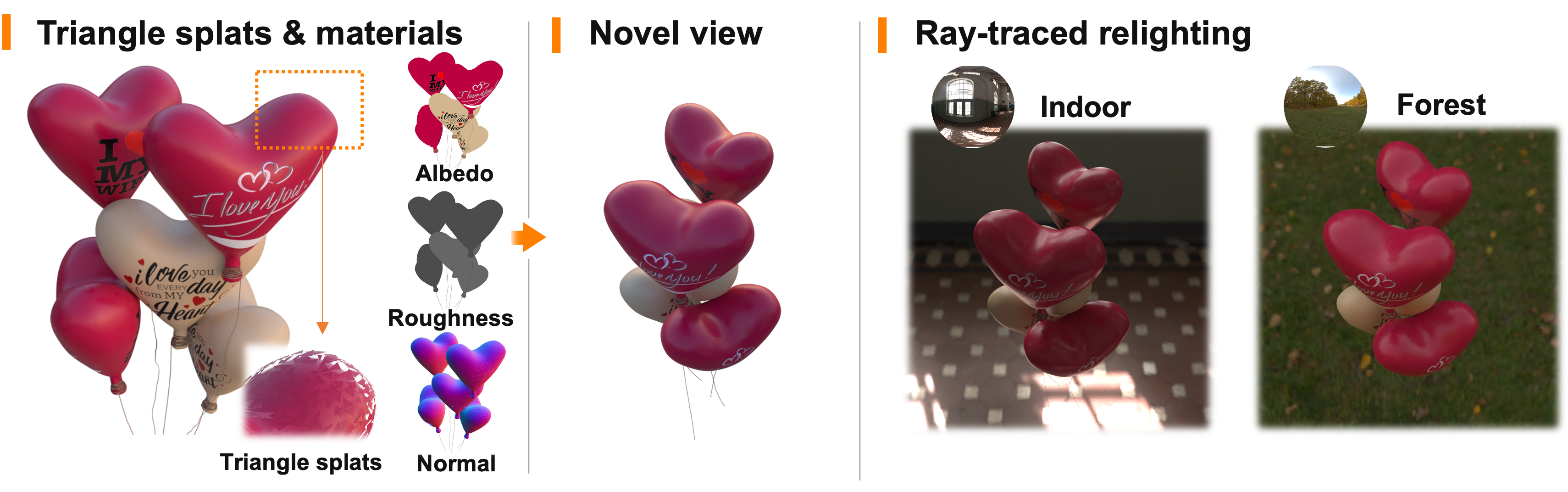}
    \captionsetup{hypcap=false}
    \captionof{figure}{We introduce TSIR, an inverse rendering method
    that reconstructs triangle splats with material attributes for ray-traced relighting
    with direct and indirect reflections.}
    \label{fig:teaser}
    \captionsetup{hypcap=true}
\end{strip}

\begin{abstract}
Inverse rendering enables realistic relighting for visual effects and augmented reality.
However, overlapping Gaussian primitives complicate surface recovery
and require costly multi-layer traversal and compositing for light transport.
We propose TSIR, a two-stage inverse rendering method
that reconstructs relightable triangle splats from multi-view images.
The first stage prepares opaque triangle splats with learnable vertex shading normals.
The second stage directly traces these surfaces to estimate materials through Monte Carlo integration
of direct illumination and indirect reflections, reducing reliance on multi-layer compositing.
Anti-aliasing improves rendering quality while maintaining efficiency.
Experiments demonstrate higher relighting quality than the evaluated baselines
and more efficient anti-aliasing than supersampling.
\end{abstract}

\section{Introduction}
\label{sec:intro}

Inverse rendering recovers geometry and materials from multi-view images,
enabling relighting and asset editing for visual effects, games, and augmented reality.
These tasks require scene representations that capture geometry and appearance
and support light transport under changing illumination.
Neural radiance fields (NeRF)~\cite{mildenhall2020nerf} provide continuous scene representations
that have been extended to inverse rendering~\cite{zhang2021nerfactor,jin2023tensoir,liu2023nero}.
However, repeated field queries along rays make optimization and relighting
computationally demanding.

To improve rendering efficiency, 3D Gaussian Splatting (3DGS)~\cite{kerbl2023gaussians}
combines explicit Gaussian primitives with fast rasterization.
GS-IR~\cite{liang2024gsir} extends this representation to inverse rendering,
using depth regularization for normals and baked volumes for occlusion and indirect illumination.
IRGS~\cite{gu2025irgs} evaluates physically based rendering (PBR) with Monte Carlo integration
and differentiable ray tracing over the planar primitives of 2D Gaussian Splatting (2DGS).
However, tracing Gaussian primitives requires traversing and compositing multiple layers,
increasing the cost of relighting.
Hybrid methods reduce this traversal cost by combining Gaussian rendering with mesh ray tracing.
GeoSplatting~\cite{ye2025geosplatting} grounds Gaussians on an optimizable mesh,
and GHPT~\cite{bo2026ghpt} combines Gaussian rasterization with mesh path tracing.
These methods use Gaussian primitives for appearance rendering and mesh surfaces for light transport,
motivating explicit surfaces that support both operations directly.

Triangle Splatting+ (TS+)~\cite{held2025triangleplus} provides such explicit surfaces,
using shared vertices and annealing to obtain opaque triangle splats.
These opaque primitives support direct ray intersections,
reducing the need for multi-layer traversal and compositing.
The same triangle surfaces support both appearance rendering and light transport.
We extend this triangle representation to PBR inverse rendering with TSIR,
a two-stage framework that reconstructs relightable triangle splats from multi-view images.

The first stage prepares triangle geometry for subsequent inverse rendering,
with triangle splats becoming opaque through annealing.
We interpolate learnable vertex shading normals to produce a smoother normal field
and suppress color patches under deferred shading.
Hybrid visibility blends per-triangle visibility during annealing
to mitigate abrupt changes near shadow boundaries.

The second stage performs inverse rendering on the reconstructed geometry,
evaluating PBR through Monte Carlo sampling of direct illumination and indirect reflections.
Surface shading combines albedo decoded from a shared triplane
with visibility and incident radiance queried by Triangle Splats Ray Tracing (TSRT).
Our anti-aliasing method suppresses artifacts at opaque triangle boundaries
while maintaining rendering efficiency.
Our contributions are:

\begin{itemize}
    \item We present TSIR, a two-stage PBR inverse rendering framework
    that directly ray traces reconstructed triangle splats to evaluate visibility
    and indirect reflections.
    \item We integrate a shared triplane albedo field into surface shading and ray tracing,
    promoting spatially consistent albedo recovery across triangle splats.
    \item We develop efficient anti-aliasing to suppress artifacts at opaque triangle boundaries.
\end{itemize}

\section{Related work}
\label{sec:related_work}

\paragraph{Scene representations.}
NeRF~\cite{mildenhall2020nerf} represents scenes as continuous density
and view-dependent radiance fields parameterized by a neural network.
NeRF-Casting~\cite{verbin2024nerfcasting} traces reflection rays through the radiance field
to model consistent reflections.
3DGS~\cite{kerbl2023gaussians} represents scenes with anisotropic Gaussians
and uses differentiable splatting with alpha blending for real-time rendering.
3D Gaussian Ray Tracing (3DGRT)~\cite{moenneloccoz2024gaussianraytracing}
enables arbitrary ray queries.
Subsequently, 3D Gaussian Unscented Transform (3DGUT)~\cite{wu2025dgut}
handles nonlinear projections and integrates with 3DGRT for secondary rays.
Gaussian Opacity Fields~\cite{yu2024opacity} improves surface extraction.
2DGS~\cite{huang2024gaussian2d} models surfaces with oriented
planar Gaussian disks and uses perspective-accurate ray-splat intersections for rendering.
IRGS~\cite{gu2025irgs} introduces differentiable 2D Gaussian Ray Tracing (2DGRT)
for visibility and indirect radiance queries.
Explicit meshes represent surfaces with connected vertices and faces
and store appearance in texture maps.
ExMesh~\cite{fan2026exmesh} directly optimizes mesh geometry and textures
with adaptive topology updates, using nvdiffrast~\cite{laine2020nvdiffrast}
for differentiable rasterization and texture sampling.
Triangle Splatting~\cite{held2026triangle} optimizes independent triangle primitives
using a differentiable window function.
TS+~\cite{held2025triangleplus} adds shared vertices, vertex color interpolation,
and an annealing schedule that produces opaque triangles.
UTrice~\cite{liu2025utrice} ray traces primitives from the first generation
of Triangle Splatting~\cite{held2026triangle},
with spherical harmonic (SH) coefficients stored per triangle.
Our tracer supports vertex attribute interpolation and surface material queries
at ray intersections for inverse rendering.

\paragraph{Inverse rendering.}
Inverse rendering recovers geometry and material properties from images
to support relighting and material editing.
Existing methods address this task using neural fields, Gaussian primitives, and explicit meshes.
NeRFactor~\cite{zhang2021nerfactor} initializes surface normals and visibility from NeRF
and estimates spatially varying reflectance with a learned
bidirectional reflectance distribution function (BRDF) prior.
TensoIR~\cite{jin2023tensoir} improves efficiency through tensor factorization
and models visibility and indirect illumination,
while NeRO~\cite{liu2023nero} separates glossy geometry reconstruction from BRDF estimation.
Explicit Gaussian rendering reduces repeated neural field evaluations,
motivating 3DGS methods such as GaussianShader~\cite{jiang2024gaussianshader},
GIR~\cite{shi2025gir}, and Relightable 3D Gaussians~\cite{gao2024relightable}.
GS-IR~\cite{liang2024gsir} improves normals through depth regularization
and uses baked occlusion to approximate light visibility for material estimation.
To improve reflective appearance, DeferredGS~\cite{wu2025deferredgs}
and GS-ROR~\cite{zhu2024gsror} evaluate shading on accumulated surface attributes.
Building on 2DGS, Ref-Gaussian~\cite{yao2025refgaussian} combines deferred PBR
with inter-reflection modeling.
IRGS~\cite{gu2025irgs} further queries visibility and indirect radiance through 2DGRT
and evaluates the rendering equation through Monte Carlo integration.
Explicit meshes also allow recovered materials to be stored as editable texture maps.
NvDiffRec~\cite{munkberg2022extracting} optimizes mesh geometry and material textures,
and NvDiffRecMC~\cite{hasselgren2022shape} adds Monte Carlo rendering and denoising.
ExMesh++~\cite{fan2026exmeshplus} first refines mesh geometry and topology,
then estimates material maps on the fixed mesh while tracing secondary rays
for one-bounce diffuse indirect illumination.
GeoSplatting~\cite{ye2025geosplatting} and GHPT~\cite{bo2026ghpt} also exploit mesh light transport
while retaining Gaussian appearance representations.
We extend TS+ to inverse rendering by estimating materials and environment illumination
on reconstructed triangle surfaces.

\begin{figure*}[t]
    \centering
    \includegraphics[width=\textwidth]{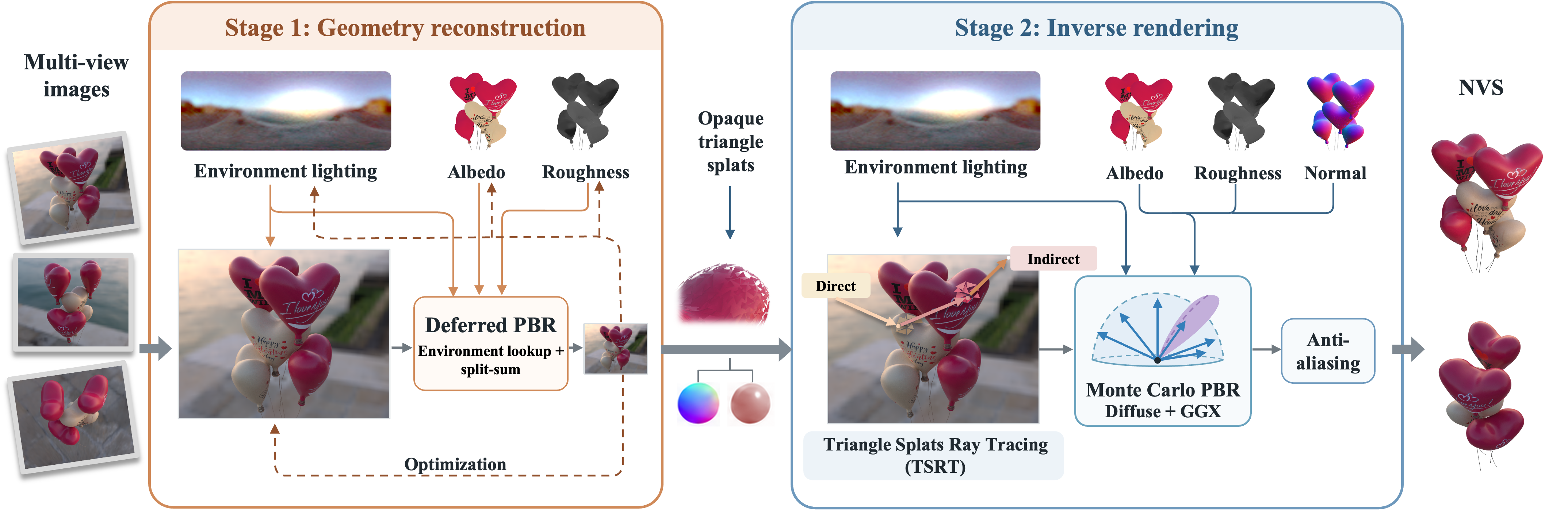}
    \caption{Overview of TSIR.
    The first stage prepares opaque triangle splats and refines vertex shading normals
    during annealing.
    The second stage refines materials through triangle ray tracing
    and Monte Carlo rendering of direct and indirect illumination.
    Anti-aliasing suppresses triangle boundary artifacts.
    The reconstructed geometry and materials form a relightable 3D asset.}
    \label{fig:pipeline}
\end{figure*}

\section{Preliminaries}
\label{sec:preliminaries}

\paragraph{Problem setting.}
Our task is to reconstruct triangle splats with material attributes
from multi-view images for relighting.
The input consists of RGB images
$\mathcal{I} = \{\mathbf{I}_k\}_{k=1}^{K}$,
where $K$ is the number of views and $\mathbf{I}_k$ is the image from view $k$.
The output asset is $\mathcal{A} = (\mathcal{V},\mathcal{F},\mathcal{M})$,
where $\mathcal{V} \subset \mathbb{R}^3$ is the shared set of vertex positions,
$\mathcal{F}$ specifies triangle connectivity through vertex indices,
and $\mathcal{M}$ denotes surface material attributes, including albedo, roughness, and metallic.
The recovered asset supports relighting by rendering its geometry and materials
under novel illumination.

\paragraph{Triangle splatting+.}
TS+~\cite{held2025triangleplus} represents a scene with shared vertices.
Each vertex stores a position $\mathbf{v}_i \in \mathbb{R}^3$,
an RGB color $\mathbf{c}_i \in \mathbb{R}^3$, and an opacity $o_i \in [0,1]$.
For a triangle $\mathcal{T}_m = (i,j,k)$ indexed by its three vertices,
the opacity is $o_m = \min(o_i,o_j,o_k)$.
The interpolated color is
\begin{equation}
    \mathbf{c}_m(\mathbf{p})
    = \sum_{v \in \{i,j,k\}} \lambda_{m,v}(\mathbf{p})\mathbf{c}_v,
    \label{eq:triangle_attributes}
\end{equation}
where $\mathbf{p} \in \mathbb{R}^2$ is a pixel position and
$\lambda_{m,v}(\mathbf{p})$ are the nonnegative barycentric coordinates
of the corresponding surface point, summing to one.
The projected triangle is defined by the maximum signed distance to its edge lines:
\begin{equation}
    \phi_m(\mathbf{p})
    = \max_{\ell \in \{1,2,3\}}
    \left(\mathbf{n}_{m\ell}^{\top}\mathbf{p} + b_{m\ell}\right),
    \label{eq:triangle_distance}
\end{equation}
where $\mathbf{n}_{m\ell} \in \mathbb{R}^2$
are the unit normals of the triangle edges pointing
outside the triangle, and $b_{m\ell}$ are offsets chosen so that
the zero-level set of $\phi_m$ defines the triangle boundary.
Thus, $\phi_m$ is positive outside the triangle, negative inside,
and zero on its boundary.
For triangle $\mathcal{T}_m$, the pixel opacity and window function are defined as
\begin{equation}
    \alpha_m(\mathbf{p}) = o_m I_m(\mathbf{p}),
    \; I_m(\mathbf{p}) = \operatorname{ReLU}\left(
        \tfrac{\phi_m(\mathbf{p})}{\phi_m(\mathbf{s}_m)}
    \right)^{\sigma},
    \label{eq:triangle_window}
\end{equation}
where $\operatorname{ReLU}(z)=\max(0,z)$ for $z \in \mathbb{R}$,
and $\mathbf{s}_m \in \mathbb{R}^2$ is the incenter of the projected triangle,
at which $\phi_m$ attains its minimum.
The shared smoothness parameter $\sigma > 0$ controls the transition
between the triangle interior and exterior.
Ordering the $M$ triangles covering $\mathbf{p}$ from front to back,
the accumulated RGB color is
\begin{equation}
    \mathbf{C}(\mathbf{p})
    = \sum_{m=1}^{M} \mathbf{c}_m(\mathbf{p})\alpha_m(\mathbf{p})
    \prod_{q=1}^{m-1}\left(1-\alpha_q(\mathbf{p})\right),
    \label{eq:triangle_compositing}
\end{equation}
where $T_m(\mathbf{p})=\prod_{q=1}^{m-1}(1-\alpha_q(\mathbf{p}))$
is the transmittance before triangle $m$.

\paragraph{Physically based rendering.}
At a non-emissive surface point $\mathbf{x} \in \mathbb{R}^3$,
the rendering equation is~\cite{pharr2023pbrt}:
\begin{equation}
    L_o(\mathbf{x},\boldsymbol{\omega}_o)
    = \int_{\Omega}\!
    f_r(\mathbf{x},\boldsymbol{\omega}_i,\boldsymbol{\omega}_o)
    L_i(\mathbf{x},\boldsymbol{\omega}_i)
    (\mathbf{n}\cdot\boldsymbol{\omega}_i)
    \,\mathrm{d}\boldsymbol{\omega}_i,
    \label{eq:rendering_equation}
\end{equation}
where $L_o$ and $L_i$ are outgoing and incident RGB radiance,
and $f_r$ is the BRDF.
The unit directions $\boldsymbol{\omega}_i$ and $\boldsymbol{\omega}_o$
point away from $\mathbf{x}$ toward the incident light and the viewer.
The integration domain $\Omega$ is the upper hemisphere
around the unit surface normal $\mathbf{n}$,
and $\mathrm{d}\boldsymbol{\omega}_i$ denotes differential solid angle.
Incident radiance includes direct illumination and light reflected by other surfaces.

Based on microfacet theory, the BRDF is defined as:
\begin{equation}
    f_r(\mathbf{x},\boldsymbol{\omega}_i,\boldsymbol{\omega}_o)
    = (1-\mu)\frac{\mathbf{a}}{\pi}
    + \frac{DFG}{
        4(\mathbf{n}\cdot\boldsymbol{\omega}_i)
        (\mathbf{n}\cdot\boldsymbol{\omega}_o)
    },
    \label{eq:microfacet_brdf}
\end{equation}
where $\mathbf{a} \in [0,1]^3$ is the linear RGB base color, termed albedo,
$r \in (0,1]$ is the roughness, and $\mu \in [0,1]$ is metallic.
Albedo is converted to sRGB for visualization.
The vector
$\mathbf{h} = (\boldsymbol{\omega}_i+\boldsymbol{\omega}_o)
/\lVert\boldsymbol{\omega}_i+\boldsymbol{\omega}_o\rVert$
is the unit half-vector.
All material quantities are evaluated at $\mathbf{x}$.
The factors $D=D(\mathbf{h},r)$,
$F=F(\boldsymbol{\omega}_o,\mathbf{h},\mathbf{F}_0)$, and
$G=G(\boldsymbol{\omega}_i,\boldsymbol{\omega}_o,r)$ denote
the microfacet normal distribution, Fresnel reflectance,
and the masking and shadowing function, respectively.
The normal-incidence reflectance is
$\mathbf{F}_0=(1-\mu)(0.04,0.04,0.04)^{\top}+\mu\mathbf{a}$,
interpolating between dielectric and metallic reflectance.

\section{Method}
\label{sec:method}

In this section, we introduce TSIR, a two-stage inverse rendering method
with triangle splats, as illustrated in \cref{fig:pipeline}.
Geometry preparation in \cref{sec:triangle_reconstruction} supplies opaque triangles
for the ray tracing and material recovery in \cref{sec:triangle_inverse_rendering}.

\subsection{Geometry reconstruction with triangle \mbox{splatting+}}
\label{sec:triangle_reconstruction}

This stage prepares opaque triangles for subsequent ray tracing and material recovery.
Training begins with forward rendering under direct illumination,
adds indirect illumination, and then switches to deferred rendering.
The TS+ annealing schedule~\cite{held2025triangleplus} decreases $\sigma$
and raises the lower bound on vertex opacity throughout training.

\paragraph{Shading normal refinement.}
TS+~\cite{held2025triangleplus} supervises face normals
using a monocular shading normal prior.
Constant face normals can produce color patches on glossy surfaces under deferred rendering.
We assign each vertex $v$ a learnable unit normal $\mathbf{n}_v^{\mathrm{s}}\in\mathbb{R}^3$.
For triangle $\mathcal{T}_m=(i,j,k)$, the shading normal at pixel $\mathbf{p}$ is
\begin{equation}
    \mathbf{n}_m^{\mathrm{s}}(\mathbf{p})
    = \frac{
        \sum_{v\in\{i,j,k\}}\lambda_{m,v}(\mathbf{p})\mathbf{n}_v^{\mathrm{s}}
    }{
        \left\lVert
            \sum_{v\in\{i,j,k\}}\lambda_{m,v}(\mathbf{p})\mathbf{n}_v^{\mathrm{s}}
        \right\rVert_2
    },
    \label{eq:shading_normal}
\end{equation}
using the barycentric coordinates from \cref{eq:triangle_attributes},
where $\lVert\cdot\rVert_2$ denotes the Euclidean norm.
In this partially connected triangle set, learned normals can fit the prior independently of geometry.
We therefore align the interpolated shading normal
with the local surface normal computed from the rendered depth map.

\paragraph{Hybrid visibility.}
Deferred rendering shades blended surface attributes.
During annealing, a single binary visibility query at the blended surface can switch abruptly
near shadow boundaries, causing sudden changes between direct and indirect specular illumination.
Specular reflection is sensitive to normals, so the optimizer may compensate
for these photometric errors through incorrect normal rotations.
We therefore blend per-triangle binary visibility using the normalized compositing weights
from \cref{eq:triangle_compositing}.
This permits a mixture of direct and indirect specular contributions as triangle weights change,
reducing the need to compensate for visibility changes by rotating normals.
Further analysis of hybrid visibility is provided in Appendix~E.

\paragraph{Training loss.}
The geometry reconstruction loss is
\begin{equation}
    \begin{aligned}
        \mathcal{L}_{\mathrm{geom}}
        &= (1-\lambda)\mathcal{L}_1
            +\lambda\mathcal{L}_{\mathrm{SSIM}}
            +\beta_1\mathcal{L}_o
            +\beta_2\mathcal{L}_n \\
        &\quad +\beta_3\mathcal{L}_{\mathrm{nc}}
            +\beta_4\mathcal{L}_d
            +\beta_5\mathcal{L}_{\mathrm{s,n}},
    \end{aligned}
    \label{eq:geometry_training_loss}
\end{equation}
where $\mathcal{L}_1$ and $\mathcal{L}_{\mathrm{SSIM}}$ denote the L1 and SSIM losses
between rendered and input images,
and $\mathcal{L}_o$ regularizes vertex opacity~\cite{held2025triangleplus}.
The prior loss $\mathcal{L}_n$ aligns the rendered normal
with the monocular shading normal prior.
The consistency loss $\mathcal{L}_{\mathrm{nc}}$ aligns the face or interpolated shading normal
with the local surface normal computed from the rendered depth map.
Both losses supervise face normals during forward rendering with indirect illumination
and shading normals during deferred rendering.
The depth distortion loss $\mathcal{L}_d$ penalizes weighted depth differences
between triangle intersections along each camera ray~\cite{huang2024gaussian2d}.
The normal smoothness loss $\mathcal{L}_{\mathrm{s,n}}$
penalizes local variations in the rendered face normal map.
The weights $\lambda\in[0,1]$ and $\beta_1,\beta_2,\beta_3,\beta_4,\beta_5\geq0$
control the loss contributions.

\subsection{Inverse rendering with triangle ray tracing}
\label{sec:triangle_inverse_rendering}

We estimate materials and environment illumination on the triangle surfaces
reconstructed in \cref{sec:triangle_reconstruction} by matching PBR renderings to the input images.
TSRT supplies visibility and incident radiance queries for Monte Carlo shading,
while a shared triplane supplies albedo consistently at rasterized and ray-traced surface points.
We then apply anti-aliasing (AA) with subpixel visibility estimation to the shaded image,
suppressing triangle boundary artifacts without additional lighting rays.

\paragraph{Triangle splats ray tracing.}
TSRT extends the triangle compositing in \cref{eq:triangle_compositing}
from projected pixels to arbitrary 3D rays.
We build a bounding volume hierarchy (BVH) directly over triangles, as in UTrice~\cite{liu2025utrice},
avoiding the icosahedral proxies used by Gaussian tracers~\cite{moenneloccoz2024gaussianraytracing,gu2025irgs}.
The shared vertex parametrization of TS+~\cite{held2025triangleplus}
supports interpolation of SH radiance within each triangle for training,
while material queries support secondary shading during relighting.

\begin{figure}[t]
    \centering
    \includegraphics[width=\linewidth]{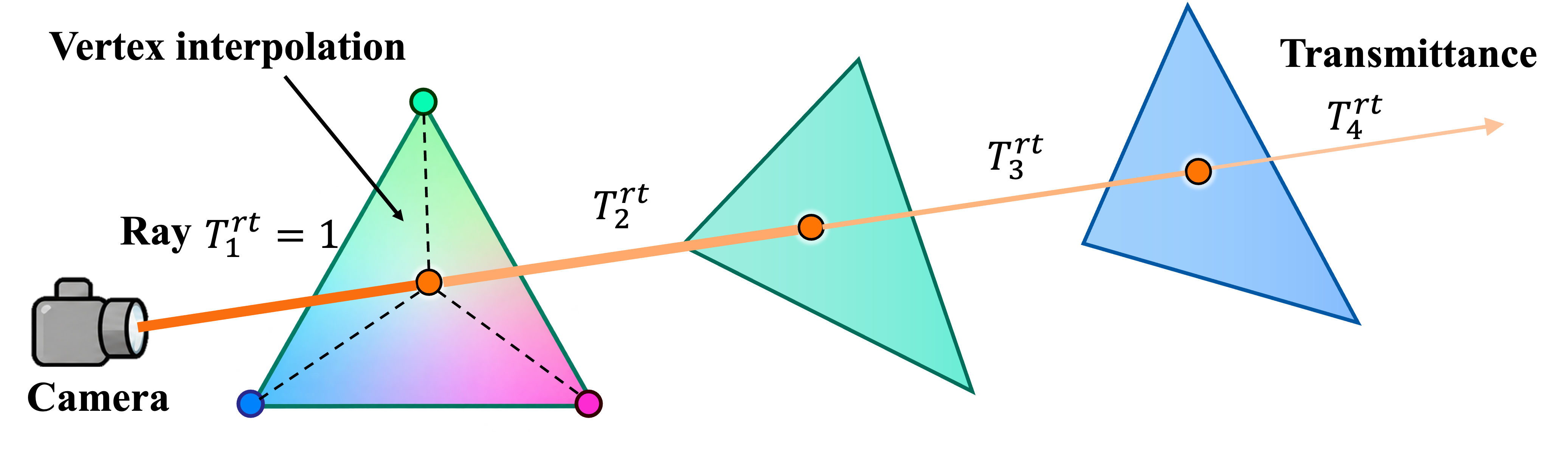}
    \caption{Triangle Splats Ray Tracing.
    Interpolated surface attributes are accumulated along each ray
    using opacity and transmittance weights.}
    \label{fig:tsrt}
\end{figure}

\Cref{fig:tsrt} illustrates how attributes are interpolated at intersections and accumulated along a ray.
For a ray with origin $\mathbf{r}_o\in\mathbb{R}^3$ and unit direction
$\mathbf{r}_d\in\mathbb{R}^3$, an intersection with $\mathcal{T}_m=(i,j,k)$ gives
\begin{equation}
    \begin{aligned}
        \mathbf{r}_o+t_m\mathbf{r}_d
        &= \sum_{a\in\{i,j,k\}}\lambda_{m,a}\mathbf{v}_a, \\
        \mathbf{c}_m^{\mathrm{rt}}
        &= \sum_{a\in\{i,j,k\}}\lambda_{m,a}\mathbf{c}_a(-\mathbf{r}_d).
    \end{aligned}
    \label{eq:tsrt_intersection}
\end{equation}
Here $t_m>0$ is the hit distance, $\lambda_{m,a}\geq0$ are barycentric coordinates summing to one,
and $\mathbf{c}_a(-\mathbf{r}_d)$ is the vertex color evaluated toward the ray origin.
As in IRGS~\cite{gu2025irgs}, each hit opacity is the primitive opacity
multiplied by its response at the intersection.
At $\mathbf{x}_m=\mathbf{r}_o+t_m\mathbf{r}_d$, we use
$\alpha_m^{\mathrm{rt}}=o_m I_m^{\mathrm{rt}}(\mathbf{x}_m)$,
where $I_m^{\mathrm{rt}}$ evaluates \cref{eq:triangle_window}
in the triangle plane using its 3D edges and incenter, with the same $o_m$ and $\sigma$.
For $N_{\mathrm{rt}}$ hits ordered from front to back, let
$T_m^{\mathrm{rt}}=\prod_{q=1}^{m-1}(1-\alpha_q^{\mathrm{rt}})$ denote the preceding transmittance.
The accumulated RGB color and opacity are
\begin{equation}
    (\mathbf{c}_{\mathrm{rt}},o_{\mathrm{rt}})
    = \sum_{m=1}^{N_{\mathrm{rt}}}T_m^{\mathrm{rt}}\alpha_m^{\mathrm{rt}}
        (\mathbf{c}_m^{\mathrm{rt}},1).
    \label{eq:tsrt_accumulation}
\end{equation}
Material attributes are accumulated with the same ray compositing weights.
The query $(\mathbf{c}_{\mathrm{rt}},o_{\mathrm{rt}},\mathbf{a}_{\mathrm{rt}},r_{\mathrm{rt}})
=\operatorname{TSRT}(\mathbf{r}_o,\mathbf{r}_d)$ returns the accumulated RGB color
$\mathbf{c}_{\mathrm{rt}}\in\mathbb{R}^3$, opacity $o_{\mathrm{rt}}\in[0,1]$,
albedo $\mathbf{a}_{\mathrm{rt}}\in[0,1]^3$, and roughness $r_{\mathrm{rt}}\in[0,1]$.
The environment visibility is $1-o_{\mathrm{rt}}$.

\paragraph{Material estimation with Monte Carlo rendering.}
We optimize materials and environment illumination by matching Monte Carlo PBR renderings
to the input images.
Deferred rendering first composites surface positions, shading normals, and materials
at each primary pixel.
We then integrate the diffuse and GGX specular responses in \cref{eq:microfacet_brdf},
replacing the environment lookup and split-sum approximation of the first stage.
For surface position $\mathbf{x}$, unit shading normal $\mathbf{n}$,
and viewing direction $\boldsymbol{\omega}_o$, the Monte Carlo estimate~\cite{gu2025irgs} is
\begin{equation}
    \mathbf{c}_{\mathrm{pbr}}
    = \frac{1}{N}\sum_{s=1}^{N}\frac{
        f_r(\mathbf{x},\boldsymbol{\omega}_s,\boldsymbol{\omega}_o)
        L_i(\mathbf{x},\boldsymbol{\omega}_s)
        (\mathbf{n}\cdot\boldsymbol{\omega}_s)
    }{p(\boldsymbol{\omega}_s)},
    \label{eq:monte_carlo_rendering}
\end{equation}
where $N$ is the number of incident directions
$\boldsymbol{\omega}_s\in\Omega$, and $p$ is their probability density with respect to solid angle.

TSRT queries visibility and indirect radiance from $\mathbf{x}$ along each sampled direction.
Following IRGS~\cite{gu2025irgs}, we combine them with environment illumination as
\begin{equation}
    L_i(\mathbf{x},\boldsymbol{\omega}_s)
    = (1-o_{\mathrm{rt}})L_{\mathrm{env}}(\boldsymbol{\omega}_s)+L_{\mathrm{ind}},
    \label{eq:incident_radiance}
\end{equation}
where $L_{\mathrm{env}}$ is environment radiance and $L_{\mathrm{ind}}$
is the accumulated indirect radiance, already weighted by opacity and transmittance.
During training, $L_{\mathrm{ind}}=\mathbf{c}_{\mathrm{rt}}$
uses the SH radiance returned by TSRT.
For relighting, we recompute one indirect bounce by shading secondary intersections
under unoccluded environment illumination using \cref{eq:monte_carlo_rendering}.

These primary and secondary BRDF evaluations use a shared triplane albedo field.
At each triangle surface point, we normalize its 3D position to the scene bounding box
and bilinearly sample features from the XY, YZ, and ZX planes.
The concatenated features are decoded into linear RGB albedo
by a shallow multilayer perceptron (MLP) with ReLU and a scaled sigmoid output.
Rasterization queries the field before pixel compositing, and TSRT queries it at ray intersections.
Sharing spatial features across nearby triangle splats promotes consistent albedo recovery
even without shared vertices.
Roughness, metallic, and shading normals retain vertex parameters.
Architecture parameters are provided in Appendix~A.3.

\begin{figure*}[!t]
    \centering
    \includegraphics[width=\textwidth]{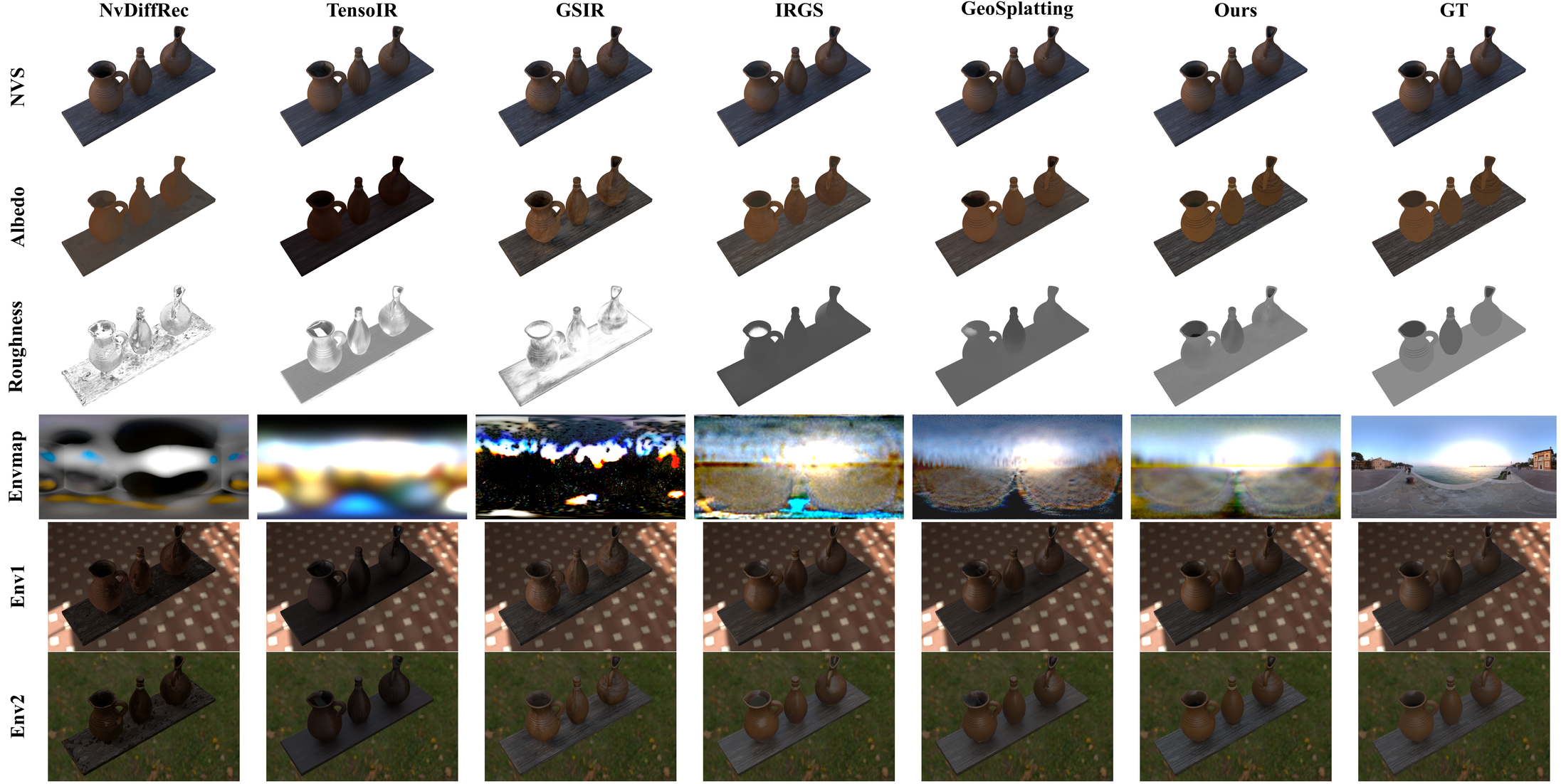}
    \caption{Qualitative comparison on the Jugs scene from Synthetic4Relight.
    Rows show novel view renderings, albedo, roughness, environment maps,
    and relighting under two novel illumination conditions.
    The rightmost column shows the ground truth.}
    \label{fig:baseline_comparison}
\end{figure*}

\paragraph{Anti-aliasing with subpixel visibility estimation.}
The Monte Carlo renderer evaluates lighting at the native pixel resolution.
Repeating this evaluation at subpixel positions, as in
supersampling anti-aliasing (SSAA)~\cite{held2025triangleplus}, increases lighting-ray cost.
Native-resolution nvdiffrast~\cite{laine2020nvdiffrast} can miss boundaries of small splats.
Our AA estimates subpixel visibility from projected geometry.

For each pixel $\mathbf{p}$, we estimate coverage by resolving the nearest triangle
at each point of a uniform $4\times4$ sample set $\mathcal{S}_p$
using projected support and intersection depth among triangles
passing projection, size, and opacity checks,
including triangles missed by the pixel center.
For a foreground sample $\mathbf{s}\in\mathcal{S}_p$,
we use the parent pixel's color if it shows the same triangle.
Otherwise, we search a $7\times7$ neighborhood for the nearest native pixel on that triangle.
If none is found, we select a foreground pixel using spatial proximity
and similarity in depth, geometric normals, and materials.
Let $\widetilde{\mathbf{P}}(\mathbf{s})\in\mathbb{R}^4$ denote its premultiplied RGBA value,
with coverage in the fourth channel.
Background samples use zero RGBA.
Aggregating the samples gives the premultiplied RGBA estimate
\begin{equation}
    \widehat{\mathbf{Q}}_p
    = \frac1{16}\sum_{\mathbf{s}\in\mathcal{S}_p}\widetilde{\mathbf{P}}(\mathbf{s}).
    \label{eq:aa_boundary_filter}
\end{equation}
After coverage normalization, we convert RGB to sRGB and composite it with the background.

\paragraph{Training loss.}
To train the indirect radiance approximation together with the material model,
we supervise SH colors and PBR renderings separately:
\begin{equation}
    \begin{aligned}
        \mathcal{L}_{\mathrm{inv}}
        &= \mathcal{L}_{\mathrm{rgb}}^{\mathrm{SH}}
            +\gamma_1\mathcal{L}_1^{\mathrm{pbr}}
            +\gamma_2\mathcal{L}_{\mathrm{light}}
            +\gamma_3\mathcal{L}_{\mathrm{s,a}} \\
        &\quad +\gamma_4\mathcal{L}_{\mathrm{s,r}}
            +\gamma_5\mathcal{L}_{\mathrm{nc}}
            +\gamma_6\mathcal{L}_{\mathrm{s,m}}
            +\gamma_7\mathcal{L}_{\mathrm{s,e}},
    \end{aligned}
    \label{eq:inverse_rendering_loss}
\end{equation}
where $\mathcal{L}_1^{\mathrm{pbr}}$ measures L1 error between the rendered PBR colors
and input images, and $\mathcal{L}_{\mathrm{light}}$ penalizes RGB channel differences
in the mean sampled incident radiance.
The independent SH color loss is
$\mathcal{L}_{\mathrm{rgb}}^{\mathrm{SH}}
= (1-\lambda)\mathcal{L}_1^{\mathrm{SH}}
+\lambda\mathcal{L}_{\mathrm{SSIM}}^{\mathrm{SH}}$.
The terms $\mathcal{L}_{\mathrm{s,a}}$, $\mathcal{L}_{\mathrm{s,r}}$,
and $\mathcal{L}_{\mathrm{s,m}}$ smooth the rendered albedo, roughness, and metallic maps
in image space, with weights determined by ground-truth image edges.
The environment smoothness loss $\mathcal{L}_{\mathrm{s,e}}$ regularizes the illumination image
obtained by querying the environment map along camera viewing directions.
The consistency loss $\mathcal{L}_{\mathrm{nc}}$ aligns the learnable shading normal
with normals derived from a fixed depth reference.
The coefficients $\gamma_1,\gamma_2,\gamma_3,\gamma_4,\gamma_5,\gamma_6,\gamma_7$
are the weights of the respective loss terms above.

\section{Experiments}
\label{sec:experiments}

\begin{table*}[t]
    \centering
    \caption{Quantitative comparison, averaged over four scenes per dataset.
    Colors mark the \colorbox{resultfirst}{best},
    \colorbox{resultsecond}{second best}, and \colorbox{resultthird}{third best}
    results.}
    \label{tab:quantitative_comparison}

    \footnotesize
    \setlength{\tabcolsep}{0.4pt}
    \begin{tabular*}{\textwidth}{@{\extracolsep{\fill}}l*{20}{c}@{}}
        \toprule
        Dataset & \multicolumn{10}{c}{Synthetic4Relight}
        & \multicolumn{10}{c}{TensoIR Synthetic} \\
        \cmidrule(lr){2-11}\cmidrule(lr){12-21}
        Method & \multicolumn{3}{c}{Novel view synthesis}
        & \multicolumn{3}{c}{Albedo} & \multicolumn{3}{c}{Relighting} & Roughness
        & \multicolumn{3}{c}{Novel view synthesis}
        & \multicolumn{3}{c}{Albedo} & \multicolumn{3}{c}{Relighting} & Normal \\
        & {\scriptsize PSNR} & {\scriptsize SSIM} & LPIPS
        & {\scriptsize PSNR} & {\scriptsize SSIM} & LPIPS
        & {\scriptsize PSNR} & {\scriptsize SSIM} & {\scriptsize LPIPS} & MSE
        & {\scriptsize PSNR} & {\scriptsize SSIM} & LPIPS
        & {\scriptsize PSNR} & {\scriptsize SSIM} & LPIPS
        & {\scriptsize PSNR} & {\scriptsize SSIM} & {\scriptsize LPIPS} & {\scriptsize MAE ($^\circ$)} \\
        \midrule
        NvDiffRec
        & 34.76 & 0.964 & 0.060
        & 31.91 & 0.940 & 0.089
        & 28.18 & 0.926 & 0.086
        & 0.0295
        & 30.13 & 0.933 & 0.082
        & 31.46 & 0.931 & 0.098
        & 26.43 & 0.893 & 0.108
        & 7.14 \\
        NvDiffRecMC
        & 33.08 & 0.950 & 0.085
        & 32.99 & 0.944 & 0.079
        & 31.19 & 0.936 & 0.092
        & 0.0082
        & 29.63 & 0.922 & 0.102
        & 33.48 & 0.950 & 0.081
        & 28.96 & 0.914 & 0.104
        & 5.94 \\
        TensoIR
        & 36.48 & \thirdscore{0.976} & 0.045
        & 31.70 & 0.939 & 0.068
        & 30.36 & 0.944 & 0.064
        & 0.0213
        & 31.42 & 0.955 & 0.052
        & \firstscore{34.86} & \thirdscore{0.951} & \secondscore{0.069}
        & \secondscore{32.03} & \secondscore{0.946} & \thirdscore{0.069}
        & 4.11 \\
        GS-IR
        & \secondscore{37.68} & \secondscore{0.977} & \secondscore{0.039}
        & 30.25 & 0.939 & 0.082
        & 30.90 & 0.956 & 0.061
        & 0.0437
        & \secondscore{36.46} & \firstscore{0.973} & \firstscore{0.035}
        & 30.98 & 0.924 & 0.098
        & 29.47 & 0.933 & \secondscore{0.065}
        & 5.34 \\
        IRGS
        & \thirdscore{37.43} & 0.975 & \thirdscore{0.040}
        & \thirdscore{33.22} & \thirdscore{0.954} & \thirdscore{0.059}
        & \secondscore{35.19} & \thirdscore{0.964} & \thirdscore{0.055}
        & \thirdscore{0.0081}
        & \thirdscore{35.51} & \thirdscore{0.964} & \thirdscore{0.049}
        & 33.92 & \secondscore{0.953} & \thirdscore{0.076}
        & \thirdscore{31.08} & \thirdscore{0.937} & 0.076
        & \secondscore{3.99} \\
        GeoSplatting
        & 35.69 & 0.975 & 0.040
        & \secondscore{34.67} & \secondscore{0.962} & \secondscore{0.050}
        & \thirdscore{34.72} & \secondscore{0.966} & \secondscore{0.052}
        & \secondscore{0.0037}
        & 33.32 & 0.957 & 0.055
        & \secondscore{34.64} & \firstscore{0.954} & \firstscore{0.069}
        & 30.80 & 0.931 & 0.076
        & \thirdscore{4.10} \\
        \midrule
        \textbf{Ours}
        & \firstscore{40.78} & \firstscore{0.982} & \firstscore{0.031}
        & \firstscore{35.10} & \firstscore{0.967} & \firstscore{0.048}
        & \firstscore{36.99} & \firstscore{0.973} & \firstscore{0.042}
        & \firstscore{0.0022}
        & \firstscore{37.10} & \secondscore{0.970} & \secondscore{0.044}
        & \thirdscore{34.43} & 0.946 & 0.078
        & \firstscore{32.75} & \firstscore{0.948} & \firstscore{0.062}
        & \firstscore{3.65} \\
        \bottomrule
    \end{tabular*}
\end{table*}

\subsection{Experimental settings}
\label{sec:experimental_settings}

\paragraph{Datasets and metrics.}
We evaluate TSIR on two synthetic datasets,
Synthetic4Relight~\cite{zhang2022invrender} and TensoIR~\cite{jin2023tensoir}.
We use PSNR, SSIM~\cite{wang2004ssim}, and LPIPS~\cite{zhang2018perceptual}
to evaluate novel view synthesis, albedo estimation, and relighting.
We evaluate rendered normal maps with mean angular error (MAE)
and roughness with mean squared error (MSE).
TSIR's normal maps use learned shading normals.
The detailed metric computations follow IRGS~\cite{gu2025irgs}.

\paragraph{Baselines.}
We compare TSIR with six inverse rendering methods.
NvDiffRec~\cite{munkberg2022extracting} optimizes mesh geometry and material textures
using differentiable rasterization.
NvDiffRecMC~\cite{hasselgren2022shape} combines Monte Carlo rendering and denoising
to jointly optimize mesh geometry, materials, and lighting.
TensoIR~\cite{jin2023tensoir} uses tensor factorization to represent radiance fields
and estimate surface reflectance.
GS-IR~\cite{liang2024gsir} equips 3D Gaussians with normal regularization
and cached occlusion to model indirect illumination.
IRGS~\cite{gu2025irgs} models inter-reflections using differentiable 2DGRT.
GeoSplatting~\cite{ye2025geosplatting} constructs Gaussians on a mesh,
using explicit surface geometry to guide light transport.

\paragraph{Implementation details.}
Implementation details are provided in Appendix~A
of the supplementary material.

\subsection{Comparison with baselines}
\label{sec:comparative_evaluation}

We conduct qualitative and quantitative comparisons of novel view synthesis,
material estimation, and relighting on Synthetic4Relight and TensoIR.
As shown in \cref{tab:quantitative_comparison},
TSIR achieves competitive results across both datasets.
On Synthetic4Relight, TSIR improves novel view PSNR by 3.10\,dB over GS-IR
and relighting PSNR by 1.79\,dB over IRGS.
On the TensoIR dataset, TSIR improves relighting PSNR by 0.73\,dB over the TensoIR baseline.
The qualitative comparison on Jugs in \cref{fig:baseline_comparison}
shows that TSIR recovers fine albedo details and accurate roughness,
and produces relighting results close to the ground truth under novel illumination.
Additional comparisons on both datasets are provided in Appendix~I
of the supplementary material.

We further evaluate relighting efficiency on TensoIR's Hotdog scene
under the bridge environment at $800\times800$ (\cref{tab:relighting_efficiency}),
using FPS as the efficiency metric, as in ExMesh++~\cite{fan2026exmeshplus}.
Timing details are provided in Appendix~D.
TSIR achieves 14.02 FPS at 31.24\,dB,
over $30\times$ the throughput of IRGS~\cite{gu2025irgs}.
Both use 768 lighting rays per pixel and shade secondary intersections for indirect illumination.
TensoIR~\cite{jin2023tensoir} samples 512 lighting directions per surface point
and uses ray marching for visibility.
NvDiffRec~\cite{munkberg2022extracting} and GS-IR~\cite{liang2024gsir}
use prefiltered environment maps with the split-sum approximation.
GS-IR renders only direct illumination in this evaluation and is faster,
while TSIR achieves 4.29\,dB higher PSNR.
GeoSplatting~\cite{ye2025geosplatting} uses 512 lighting rays per pixel
and approximates indirect radiance without PBR evaluation at secondary intersections,
contributing to its higher throughput.
The sampling ablation in \cref{sec:ablation_studies,tab:ablation_efficiency}
shows that reducing TSIR to 192 rays achieves 35.17 FPS at 30.76\,dB,
comparable in speed to GeoSplatting's 32.03 FPS at 29.21\,dB, with higher PSNR.

\begin{table}[!t]
    \centering
    \caption{Relighting efficiency on TensoIR's Hotdog scene under the bridge environment
    at $800\times800$.}
    \label{tab:relighting_efficiency}
    \small
    \setlength{\tabcolsep}{3.5pt}

    \begin{tabular*}{\columnwidth}{@{\extracolsep{\fill}}l*{5}{c}@{}}
        \toprule
        & \#Primitives & FPS$\uparrow$
        & PSNR$\uparrow$ & SSIM$\uparrow$ & LPIPS$\downarrow$ \\
        \midrule
        NvDiffRec
        & 24k & 232.91 & 20.37 & 0.8666 & 0.1543 \\
        NvDiffRecMC
        & 26k & 10.83 & 26.29 & 0.9017 & 0.1292 \\
        TensoIR
        & -- & 0.05 & 25.68 & 0.9240 & 0.1071 \\
        GS-IR
        & 139k & 113.22 & 26.95 & 0.9234 & 0.0878 \\
        IRGS
        & 38k & 0.46 & 29.18 & 0.9155 & 0.1174 \\
        GeoSplatting
        & 280k & 32.03 & 29.21 & 0.9399 & 0.0958 \\
        \textbf{Ours}
        & 67k & 14.02 & 31.24 & 0.9372 & 0.0974 \\
        \bottomrule
    \end{tabular*}
\end{table}

\subsection{Ablation studies}
\label{sec:ablation_studies}

To evaluate the effectiveness of TSIR's components, we conduct ablation studies
on shading normals, triplane albedo, anti-aliasing, and lighting sample counts.
\Cref{tab:ablation_studies} summarizes the quality results on TensoIR,
while \cref{tab:ablation_efficiency} compares the efficiency of anti-aliasing and sampling variants
on Hotdog.
Additional ablation results are provided in the supplementary material.

\begin{table*}[!t]
    \centering
    \caption{Ablations on TensoIR, averaged equally over four scenes.
    Relighting averages five environments per scene.
    w/o indirect disables indirect radiance only at inference.}
    \label{tab:ablation_studies}
    \fontsize{9}{10}\selectfont
    \setlength{\tabcolsep}{0.5pt}

    \begin{tabular*}{\textwidth}{@{\extracolsep{\fill}}l*{10}{c}@{}}
        \toprule
        & \multicolumn{3}{c}{Novel view synthesis} & \multicolumn{3}{c}{Albedo}
        & \multicolumn{3}{c}{Relighting} & Normal \\
        \cmidrule(lr){2-4}\cmidrule(lr){5-7}\cmidrule(lr){8-10}
        & PSNR
        & SSIM
        & LPIPS
        & PSNR
        & SSIM
        & LPIPS
        & PSNR
        & SSIM
        & LPIPS
        & MAE$\downarrow$ \\
        \midrule
        \textbf{Ours}
        & 37.10 & 0.9703 & 0.0444 & 34.43 & 0.9457 & 0.0784
        & 32.75 & 0.9484 & 0.0621 & 3.653 \\
        \midrule
        \multicolumn{11}{@{}l}{\textit{(a) Shading normal}} \\
        Face normal
        & 36.43 & 0.9674 & 0.0474 & 33.79 & 0.9413 & 0.0852
        & 32.36 & 0.9459 & 0.0644 & 3.589 \\
        \midrule
        \multicolumn{11}{@{}l}{\textit{(b) Triplane albedo}} \\
        Vertex albedo
        & 36.88 & 0.9672 & 0.0509 & 33.89 & 0.9469 & 0.0780
        & 32.36 & 0.9462 & 0.0674 & 3.715 \\
        DiffSoup
        & 34.71 & 0.9480 & 0.0715 & 32.80 & 0.9155 & 0.1161
        & 31.10 & 0.9286 & 0.0826 & 4.084 \\
        \midrule
        \multicolumn{11}{@{}l}{\textit{(c) Anti-aliasing}} \\
        w/o AA
        & 34.10 & 0.9593 & 0.0651 & 34.07 & 0.9409 & 0.0862
        & 31.67 & 0.9380 & 0.0786 & 3.914 \\
        SSAA
        & 37.18 & 0.9725 & 0.0371 & 34.40 & 0.9459 & 0.0777
        & 32.90 & 0.9518 & 0.0565 & 3.642 \\
        nvdiffrast AA
        & 35.73 & 0.9657 & 0.0591 & 34.25 & 0.9433 & 0.0843
        & 32.42 & 0.9440 & 0.0720 & 3.761 \\
        \midrule
        \multicolumn{11}{@{}l}{\textit{(d) PBR rendering}} \\
        w/o indirect
        & 34.27 & 0.9655 & 0.0499 & 34.43 & 0.9457 & 0.0784
        & 32.45 & 0.9477 & 0.0650 & 3.653 \\
        spp=384
        & 37.01 & 0.9685 & 0.0489 & 34.43 & 0.9457 & 0.0784
        & 32.68 & 0.9449 & 0.0669 & 3.653 \\
        spp=192
        & 36.83 & 0.9651 & 0.0539 & 34.43 & 0.9457 & 0.0784
        & 32.55 & 0.9390 & 0.0739 & 3.653 \\
        spp=96
        & 36.50 & 0.9591 & 0.0608 & 34.43 & 0.9457 & 0.0784
        & 32.30 & 0.9302 & 0.0835 & 3.653 \\
        \bottomrule
    \end{tabular*}
\end{table*}

\begin{figure}[!htb]
    \centering
    \includegraphics[width=\linewidth]{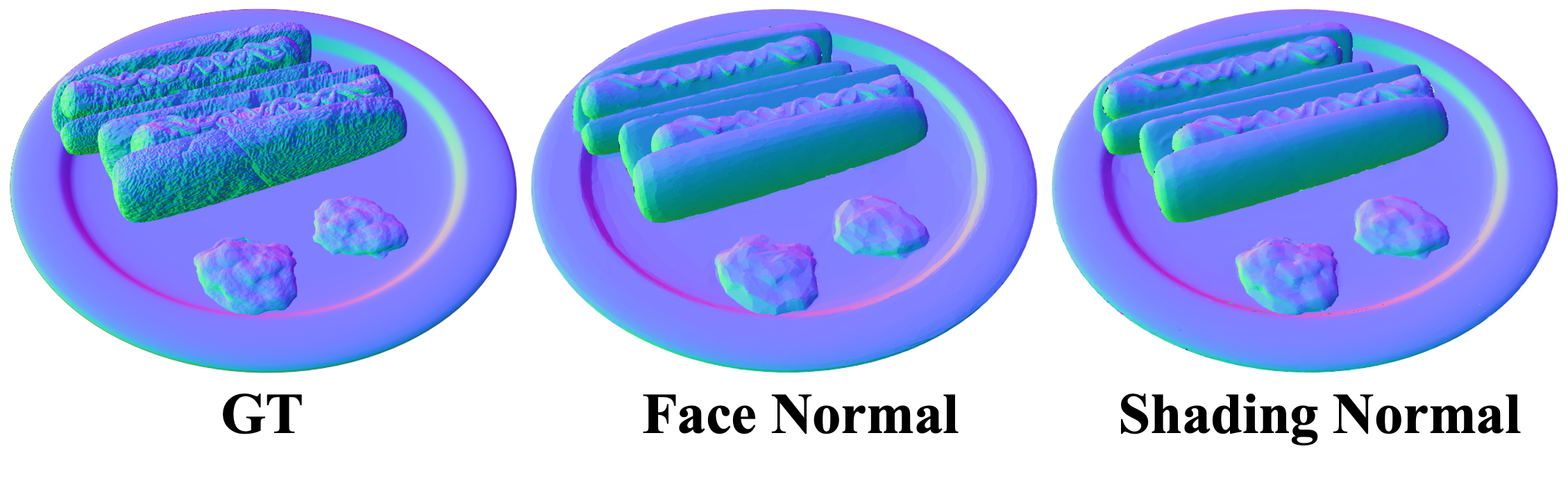}
    \caption{Shading normal ablation on the Hotdog scene.
    From left to right: ground truth, face normals, and shading normals.}
    \label{fig:ablation_shading_normal}
\end{figure}

\paragraph{Shading normal.}
\Cref{tab:ablation_studies} shows that shading normals do not necessarily
reduce rendered normal MAE,
but improve the fit of the PBR rendering equation to the observed images.
They improve novel view and relighting PSNR by 0.67 and 0.39\,dB, respectively.
As shown in \cref{fig:ablation_shading_normal}, shading normals produce smoother rendered normal maps
than face normals, alleviating normal discontinuities as triangle splats become opaque.

\begin{figure}[!htb]
    \centering
    \includegraphics[width=\linewidth]{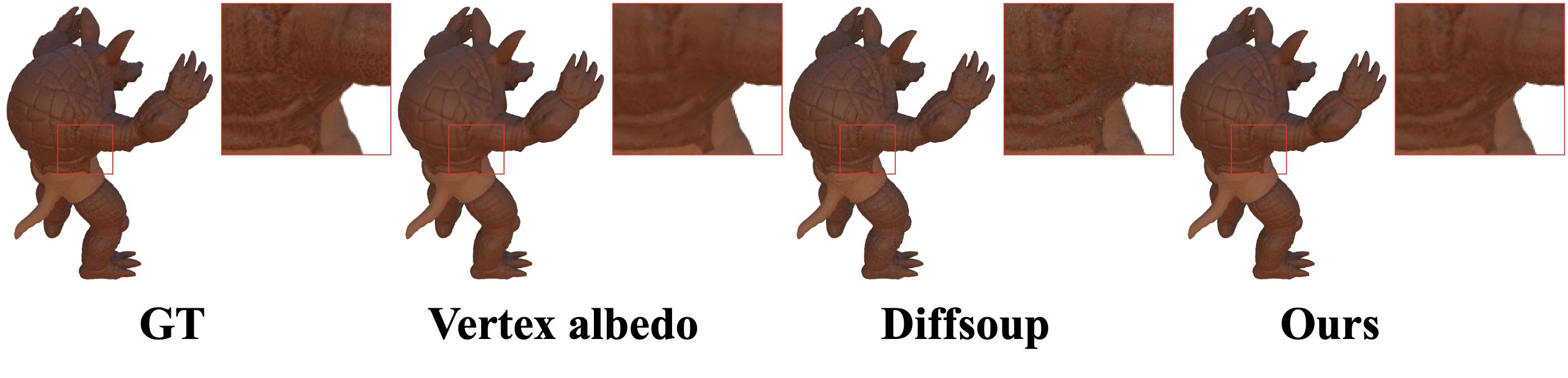}
    \caption{Triplane albedo ablation on the Armadillo scene.
    Insets highlight local albedo details.}
    \label{fig:ablation_triplane}
\end{figure}

\paragraph{Triplane albedo.}
\Cref{tab:ablation_studies} shows that triplane albedo improves albedo PSNR
by 0.53\,dB over vertex albedo and 1.63\,dB over DiffSoup,
with corresponding gains in novel view and relighting quality.
DiffSoup~\cite{tojo2026diffsoup} increases the representational capacity of triangle splats
through per-triangle textures.
In our inverse rendering setting, these textures tend to overfit when optimized from scratch.
\Cref{fig:ablation_triplane} shows that our method recovers richer albedo details
on the back of Armadillo.
Further analysis of triplane albedo is provided in Appendix~F
of the supplementary material.

\begin{figure}[!htb]
    \centering
    \includegraphics[width=\linewidth]{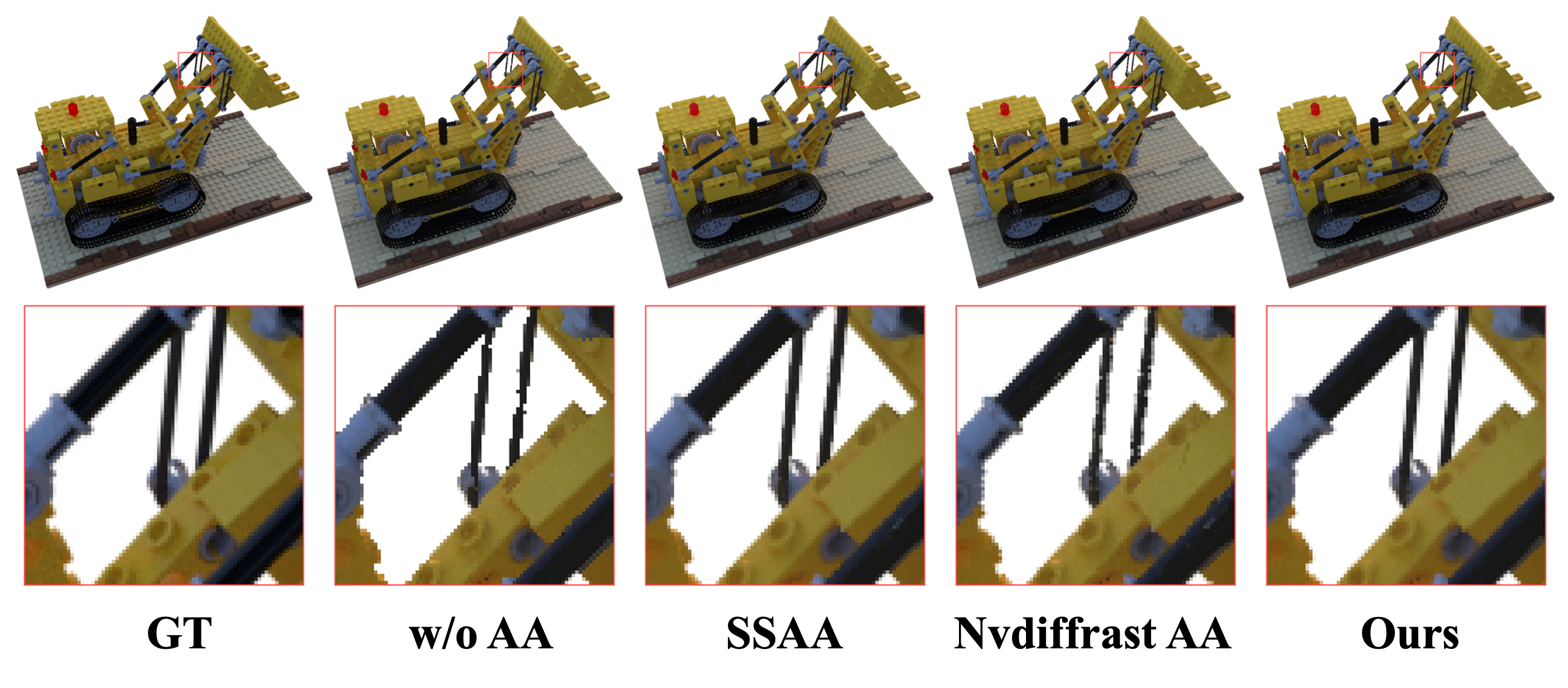}
    \caption{Anti-aliasing ablation on the Lego scene.
    Insets compare boundary details.}
    \label{fig:ablation_aa}
\end{figure}

\paragraph{Anti-aliasing.}
We compare our AA with no AA, SSAA, and nvdiffrast AA.
\Cref{tab:ablation_studies} shows that our filter achieves quality comparable to SSAA
and outperforms the other two variants, with novel view PSNR of 37.10\,dB
versus 37.18\,dB for SSAA.
\Cref{fig:ablation_aa} compares the resulting boundary details.
SSAA evaluates PBR at $1600\times1600$ before downsampling to $800\times800$,
while our method evaluates PBR directly at $800\times800$.
By shading fewer pixels, our method achieves 13.81 FPS on Hotdog,
$3.52\times$ the throughput of SSAA, with only 0.16\,dB lower relighting PSNR
(\cref{tab:ablation_efficiency}).
Additional ablations are provided in Appendix~G.

\paragraph{PBR rendering.}
We evaluate the effects of indirect illumination and lighting sample counts
on rendering quality and efficiency.
Removing indirect illumination increases Hotdog throughput from 14.02 to 17.45 FPS,
but reduces relighting PSNR from 31.24 to 28.87\,dB (\cref{tab:ablation_efficiency}).
We reduce the number of lighting rays per pixel from 768 to 384, 192, and 96.
\Cref{tab:ablation_efficiency} shows that at 96 rays, Hotdog throughput increases
from 14.02 to 49.56 FPS, while relighting PSNR decreases from 31.24 to 30.19\,dB.
The average PSNR drops on TensoIR are 0.60\,dB for novel view synthesis
and 0.45\,dB for relighting (\cref{tab:ablation_studies}).

\begin{table}[!t]
    \centering
    \caption{Relighting efficiency ablations on Hotdog under the bridge environment.
    All AA variants use 768 samples per pixel (spp).
    w/o indirect disables indirect radiance only at inference.}
    \label{tab:ablation_efficiency}
    \small

    \begin{tabular*}{\columnwidth}{@{\extracolsep{\fill}}l*{4}{c}@{}}
        \toprule
        & FPS$\uparrow$
        & PSNR$\uparrow$ & SSIM$\uparrow$ & LPIPS$\downarrow$ \\
        \midrule
        \multicolumn{5}{@{}l}{\textit{(a) Anti-aliasing}} \\
        w/o AA
        & 13.93 & 30.22 & 0.9268 & 0.1137 \\
        SSAA
        & 3.92 & 31.39 & 0.9496 & 0.0846 \\
        nvdiffrast, AA
        & 13.15 & 30.80 & 0.9313 & 0.1077 \\
        \textbf{Ours}$^*$
        & 13.81 & 31.24 & 0.9372 & 0.0974 \\
        \midrule
        \multicolumn{5}{@{}l}{\textit{(b) PBR rendering}} \\
        \textbf{Ours}
        & 14.02 & 31.24 & 0.9372 & 0.0974 \\
        w/o indirect
        & 17.45 & 28.87 & 0.9300 & 0.1041 \\
        spp=384
        & 22.22 & 31.07 & 0.9236 & 0.1054 \\
        spp=192
        & 35.17 & 30.76 & 0.9035 & 0.1158 \\
        spp=96
        & 49.56 & 30.19 & 0.8778 & 0.1293 \\
        \bottomrule
    \end{tabular*}
    \par\smallskip
    \begin{minipage}{\columnwidth}
        \footnotesize
        $^*$To accommodate the higher GPU memory requirements of SSAA,
        all variants in (a) sample pixels in two batches.
    \end{minipage}
\end{table}

\section{Conclusion}
\label{sec:conclusion}

We presented TSIR, a two-stage inverse rendering method
that reconstructs relightable triangle splats from multi-view images.
An initial reconstruction stage prepares opaque triangles for ray tracing.
The second stage integrates TSRT with Monte Carlo rendering to model direct illumination
and indirect reflections for material estimation.
Tracing opaque triangles reduces multi-layer alpha blending overhead,
and anti-aliasing suppresses artifacts at triangle boundaries.
Experiments demonstrate that TSIR achieves high rendering quality and computational efficiency.

{
    \small
    \bibliographystyle{ieeenat_fullname}
    \bibliography{main}
}

\clearpage
\maketitlesupplementary
\appendix

This supplementary material provides additional implementation details,
qualitative results on Stanford-ORB, a comparison with TS+ in the first stage,
a comparison of soft and opaque triangle splats, an ablation of hybrid visibility,
an analysis of triplane albedo,
and additional results on TensoIR and Synthetic4Relight.

\section{Additional implementation details}
\label{sec:supp_implementation}

\subsection{Data preparation and evaluation}
\label{sec:supp_data_evaluation}
We follow the original training and test splits of TensoIR~\cite{jin2023tensoir}
and Synthetic4Relight~\cite{zhang2022invrender}.
Each baseline is retrained on the corresponding training split.
We obtain normal priors using the same model as TS+~\cite{held2025triangleplus}.

\subsection{Optimization settings}
\label{sec:supp_optimization}
All experiments use an NVIDIA RTX 4090 GPU.
Geometry reconstruction runs for 30,000 iterations: forward rendering with direct illumination only
at iterations 1 to 5,000, adding indirect illumination at 5,001 to 10,000,
then deferred rendering with hybrid visibility after 10k iterations.
Triangle smoothness and the vertex opacity floor follow the annealing schedule
of TS+~\cite{held2025triangleplus}.
In \cref{eq:geometry_training_loss}, $\lambda=0.2$ throughout training.
Regularization uses $\beta_1=1.9\times10^{-6}$ only at iterations 1 to 4,999,
$\beta_2=\beta_3=0.05$ and $\beta_4=10$ from iteration 5,001 onward,
and $\beta_5=0.05$ only at iterations 5,001 to 10,000.
Monocular and consistency supervision switches from face to shading normals after 10k iterations.
For geometry reconstruction, we adopt the indirect radiance and mesh visibility
implementation of Ref-Gaussian~\cite{yao2025refgaussian}.
Inverse rendering uses Adam and deferred rendering for 20,000 iterations,
with weights in \cref{eq:inverse_rendering_loss} set to $\gamma_1=1$,
$\gamma_2=0.01$, $\gamma_3=\gamma_4=\gamma_6=0.2$,
$\gamma_5=0.05$, and $\gamma_7=0.0005$.

The learning rates are shared across all scenes.
Both stages use $0.0075$ for albedo,
$0.005$ for vertex roughness and shading-normal parameters,
$0.01$ for the environment map,
and $0.0075$ and $0.000375$ for the constant term and remaining vertex SH coefficients,
respectively.
Albedo uses vertex parameters in the first stage and triplane features and the decoder
in the second stage.
The SH rates apply to both appearance and indirect radiance in the first stage.
The first-stage specular strength and second-stage metallic parameters both use $0.005$.
In the first stage, vertex opacity uses $0.03$, and shading-normal optimization starts
at iteration 10,001.
Vertex positions use an exponential learning-rate schedule from $0.0015$ to $0.000015$
over the 30,000 iterations of the first stage, with a zero rate for the first 999 iterations.

\subsection{Triplane albedo architecture}
\label{sec:supp_triplane_architecture}
The albedo field uses XY, YZ, and ZX feature planes,
each with a $512\times512$ grid and eight channels.
Positions are normalized to $[-1,1]$ using the bounding box of the first-stage geometry,
expanded by 1\% at both ends of each axis.
Bilinear sampling uses aligned grid corners and border padding.
The three sampled feature vectors are concatenated into a 24-dimensional input.
The MLP uses fully connected layers of dimensions $24\to32\to3$,
with ReLU between the layers and a sigmoid output rescaled to $[0.01,0.99]$ in linear RGB.
Plane features are initialized from a zero-mean Gaussian distribution with standard deviation $0.01$.
The final decoder layer has zero weights and a bias that reproduces the inherited constant albedo.
The planes and decoder are optimized jointly with the albedo learning rate
in \cref{sec:supp_optimization}.

\subsection{Rendering details}
\label{sec:supp_rendering}
Inverse rendering training uses 128 uniform hemisphere samples and 128 samples
from the GGX distribution of visible normals (VNDF) per pixel.
Relighting uses 256 uniform, 256 GGX VNDF,
and 256 environment map importance samples per pixel.
Both use multiple importance sampling (MIS).
During training, TSRT queries visibility and accumulates vertex SH colors
to approximate indirect radiance.
Relighting evaluates one indirect bounce, with secondary surfaces shaded
by unoccluded environment illumination.

\section{Qualitative results on Stanford-ORB}
\label{sec:supp_stanford_orb}

\begin{figure}[!t]
    \centering
    \includegraphics[width=\columnwidth]{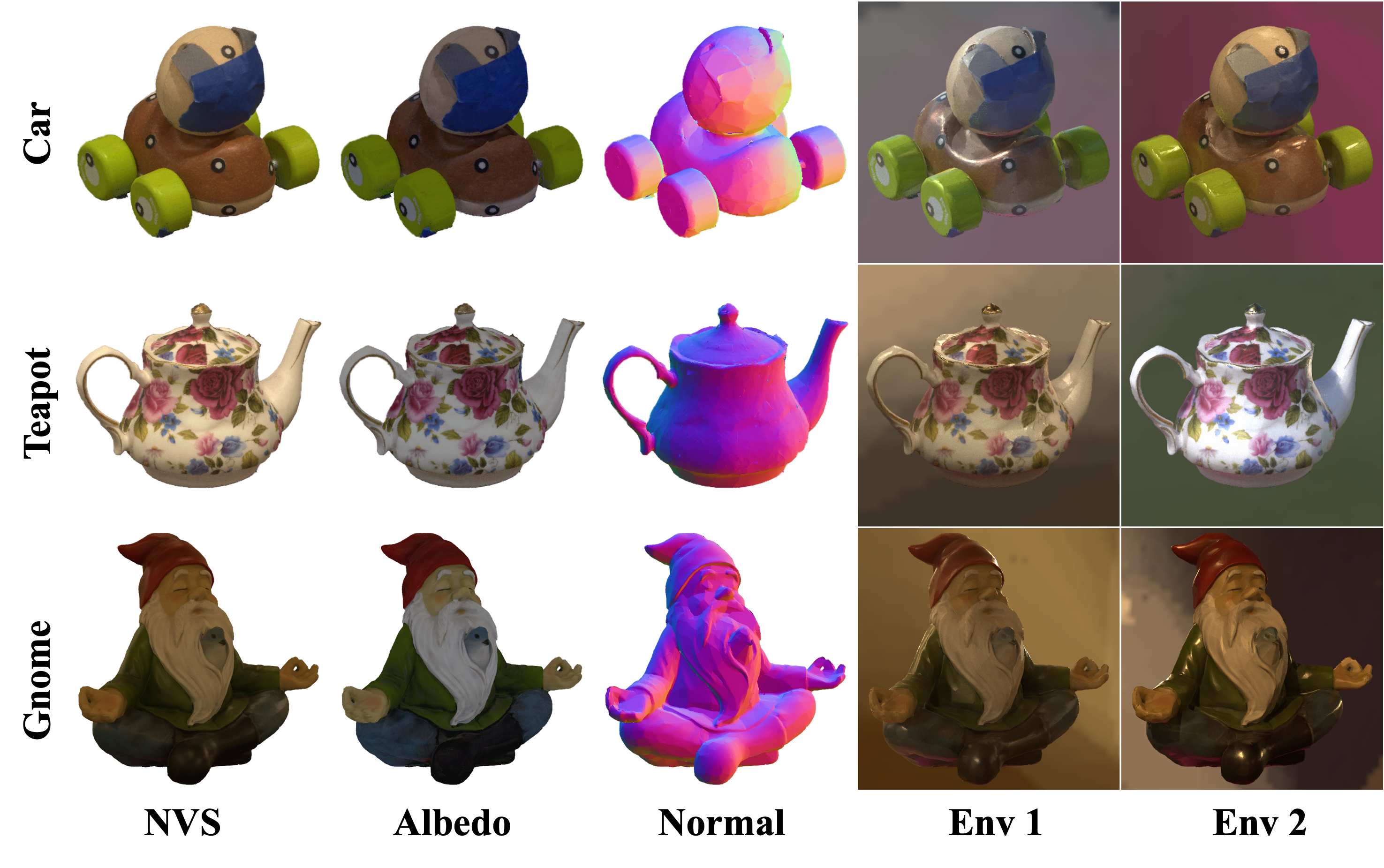}
    \caption{Qualitative results of TSIR on the real-world Stanford-ORB dataset.}
    \label{fig:supp_stanford_orb}
\end{figure}

We provide qualitative results on the real-world Stanford-ORB dataset
in \cref{fig:supp_stanford_orb}.
These include novel view renderings, estimated albedo, rendered shading normals,
and relighting under two environment maps.

\section{First-stage comparison with TS+}
\label{sec:supp_stage1_comparison}

\subsection{Experimental settings}
\label{sec:supp_stage1_settings}
We compare TS+ and the first stage of TSIR on the four TensoIR scenes.
The comparison focuses on rendered normal maps and novel view rendering.
TSIR uses its learned shading normals.
Both methods use the same data splits and are evaluated after 30,000 iterations,
retaining their respective training settings.
TS+ uses the random background setting.
Both methods rasterize normal maps at $3200\times3200$ using $4\times$ SSAA along each image axis
and apply area downsampling to $800\times800$ before evaluation.
Normal MAE is computed over the full image using rendered normals in world coordinates.

\subsection{Results}
\label{sec:supp_stage1_results}
\begin{figure}[!ht]
    \centering
    \includegraphics[width=\columnwidth]{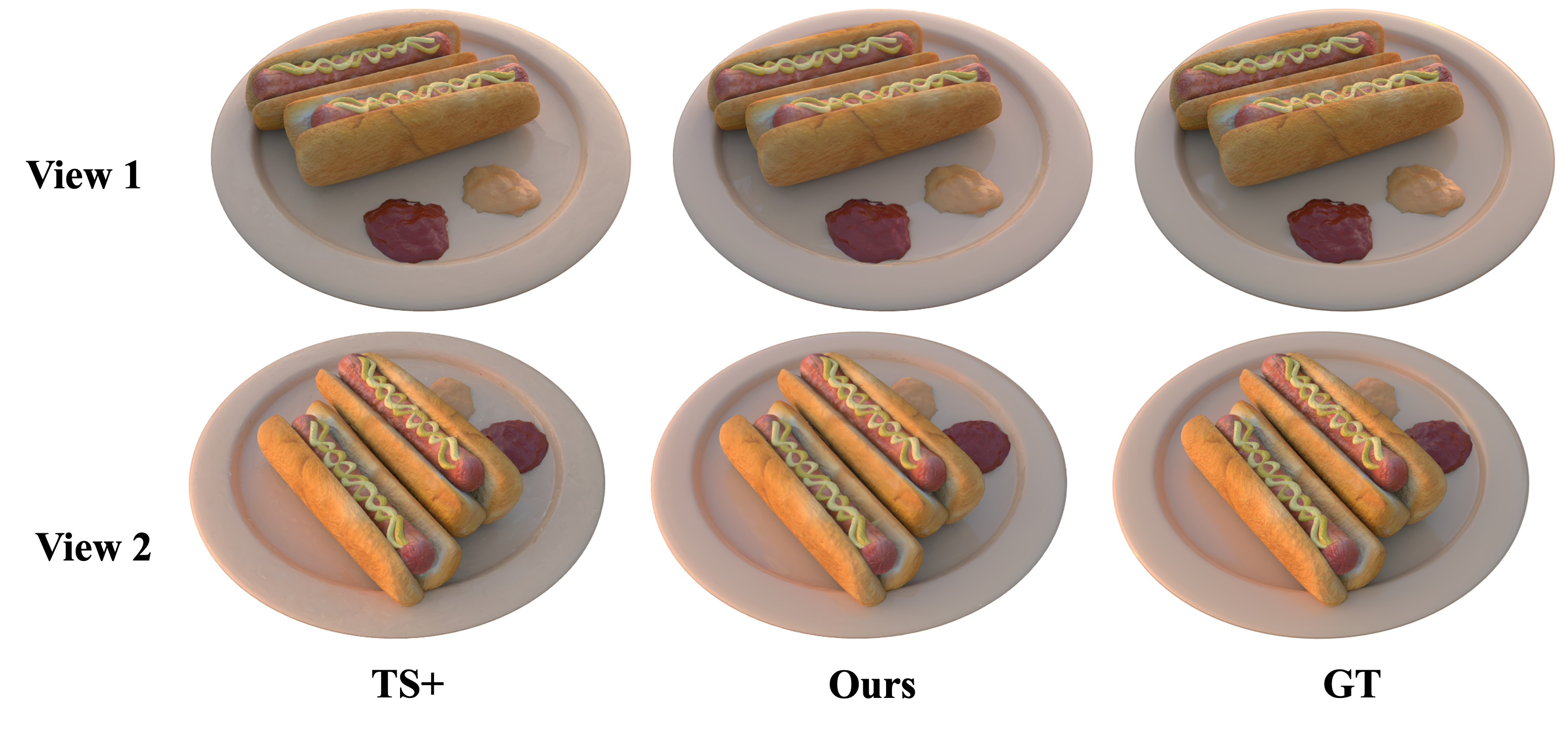}
    \caption{First-stage renderings of Hotdog.
    Columns show TS+, TSIR, and ground truth.}
    \label{fig:supp_stage1_renderings}
\end{figure}

\begin{figure*}[!t]
    \centering
    \includegraphics[width=\textwidth]{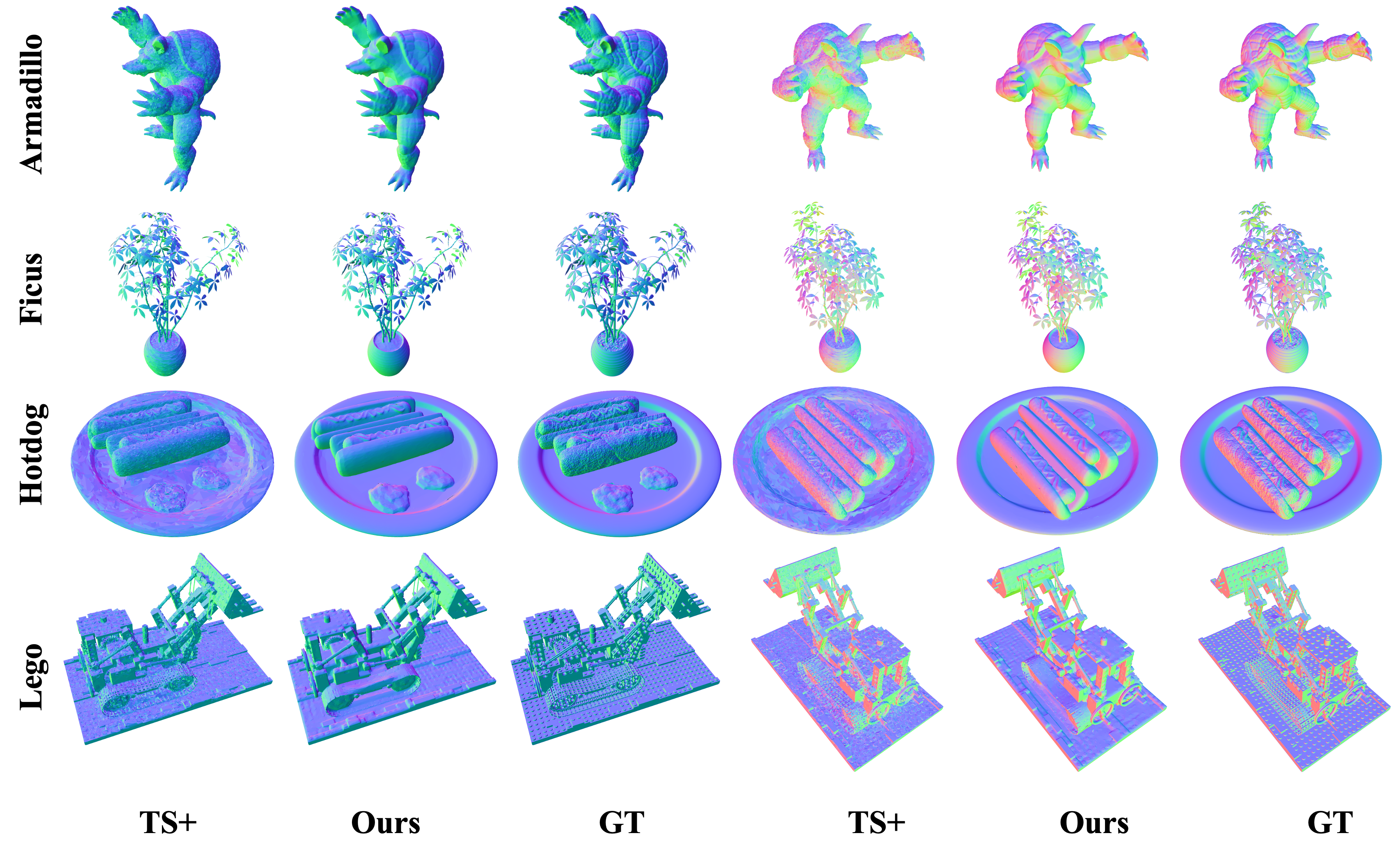}
    \caption{First-stage rendered normal comparison on TensoIR.
    Rows show Armadillo, Ficus, Hotdog, and Lego.
    Each group of three columns shows TS+, TSIR, and ground truth for one view.}
    \label{fig:supp_stage1_normals}
\end{figure*}

\Cref{tab:supp_stage1_comparison} reports full-image rendered normal MAE.
TSIR reduces the four-scene average from $4.397^\circ$ to $3.547^\circ$,
with lower errors on Armadillo, Ficus, and Hotdog.
On Lego, TS+ achieves $6.733^\circ$, compared with $7.115^\circ$ for TSIR.
On the glossy Hotdog scene, TSIR produces more faithful novel view renderings than TS+,
as shown in \cref{fig:supp_stage1_renderings}.
It is worth noting that, under strong geometric constraints, TSIR may not always
achieve higher rendering quality than TS+ on non-glossy objects.
\Cref{fig:supp_stage1_normals} compares the rendered normal maps.

\begin{table}[!ht]
    \centering
    \caption{First-stage rendered normal comparison on TensoIR with matched SSAA.
    Values are full-image rendered normal MAE ($^\circ$)$\downarrow$.
    Average denotes the equal mean over four scenes.}
    \label{tab:supp_stage1_comparison}
    \small

    \begin{tabular*}{\columnwidth}{@{\extracolsep{\fill}}l*{5}{c}@{}}
        \toprule
        Method & Armadillo & Ficus & Hotdog & Lego & Average \\
        \midrule
        TS+ & 2.785 & 3.067 & 5.001 & \textbf{6.733} & 4.397 \\
        Ours & \textbf{1.588} & \textbf{2.624} & \textbf{2.861}
        & 7.115 & \textbf{3.547} \\
        \bottomrule
    \end{tabular*}
\end{table}

\section{Analysis of opaque triangle splats}
\label{sec:supp_soft_geometry}

We analyze the quality and efficiency of opaque triangle splats
by comparing them with soft triangle splats at matched primitive counts
on the four Synthetic4Relight scenes.
Each primitive is a triangle splat, with counts listed in \cref{tab:supp_soft_geometry_quality}.
The soft variant uses $\sigma=1$, a zero vertex opacity floor, and no AA.
The opaque variant uses annealing and our AA.
All variants train the second stage for 20,000 iterations.
Quality follows the metrics in \cref{sec:experimental_settings},
with relighting averaged over envmap6 and envmap12.

\Cref{tab:supp_soft_geometry_quality} shows that Soft achieves higher average image quality,
with modest gains over Opaque.
Average novel view PSNR increases from 40.78 to 41.92\,dB,
while roughness MSE increases from 0.0022 to 0.0025.

For efficiency, we compare Opaque and Soft at matched primitive counts
under envmap6 (\cref{tab:supp_soft_geometry}).
Relighting uses $800\times800$ images, 768 lighting rays per pixel, MIS,
and indirect illumination on an NVIDIA RTX 4090 GPU.
Timing includes acceleration structure construction and environment preprocessing.
The configurations retain their own learned geometry and attributes,
so the measured speedups reflect the complete rendering pipelines.
The opaque variant reaches 17.87 FPS overall, compared with 1.16 FPS for soft splats,
corresponding to a $15.44\times$ speedup.
The modest image-quality gains of Soft therefore come at a substantial cost in relighting efficiency.

\begin{table}[!t]
    \centering
    \caption{Relighting efficiency of Soft and Opaque under envmap6.
    Both configurations use the same number of primitives per scene, rounded to the nearest thousand.
    Overall reports the mean primitive count and computes FPS from total elapsed time across scenes.}
    \label{tab:supp_soft_geometry}
    \small
    \setlength{\tabcolsep}{4pt}
    \begin{tabular*}{\columnwidth}{@{\extracolsep{\fill}}lcccc@{}}
        \toprule
        Scene & \#Primitives & \multicolumn{2}{c}{FPS$\uparrow$} & Speedup \\
        \cmidrule(lr){3-4}
        & & Soft & Opaque & \\
        \midrule
        Air balloons & 82k & 1.20 & 16.87 & $14.09\times$ \\
        Chair & 65k & 1.24 & 21.03 & $16.97\times$ \\
        Hotdog & 67k & 0.84 & 13.85 & $16.49\times$ \\
        Jugs & 46k & 1.60 & 22.29 & $13.92\times$ \\
        \midrule
        Overall & 65k & 1.16 & 17.87 & $15.44\times$ \\
        \bottomrule
    \end{tabular*}
\end{table}

\begin{table*}[!t]
    \centering
    \caption{Quality comparison of soft and opaque triangle splats on Synthetic4Relight.
    Soft and Opaque use the same number of primitives in each scene.
    The number of primitives is rounded to the nearest thousand.
    Relighting averages envmap6 and envmap12. Average gives the equal mean over four scenes.}
    \label{tab:supp_soft_geometry_quality}
    \small
    \setlength{\tabcolsep}{1.5pt}
    \begin{tabular*}{\textwidth}{@{\extracolsep{\fill}}llc*{10}{c}@{}}
        \toprule
        Scene & Variant & \#Primitives & \multicolumn{3}{c}{Novel view synthesis}
        & \multicolumn{3}{c}{Albedo} & Roughness & \multicolumn{3}{c}{Relighting} \\
        \cmidrule(lr){4-6}\cmidrule(lr){7-9}\cmidrule(lr){11-13}
        & & & PSNR$\uparrow$ & SSIM$\uparrow$ & LPIPS$\downarrow$
        & PSNR$\uparrow$ & SSIM$\uparrow$ & LPIPS$\downarrow$ & MSE$\downarrow$
        & PSNR$\uparrow$ & SSIM$\uparrow$ & LPIPS$\downarrow$ \\
        \midrule
        Air balloons & Opaque & 82k
        & 37.38 & 0.9741 & 0.0351 & 33.64 & 0.9555 & 0.0408
        & 0.0008 & 36.48 & 0.9643 & 0.0459 \\
         & Soft & 82k
        & 39.61 & 0.9783 & 0.0260 & 34.15 & 0.9588 & 0.0345
        & 0.0012 & 36.59 & 0.9652 & 0.0374 \\
        \midrule
        Chair & Opaque & 65k
        & 41.56 & 0.9825 & 0.0158 & 32.87 & 0.9718 & 0.0311
        & 0.0029 & 37.01 & 0.9760 & 0.0214 \\
         & Soft & 65k
        & 42.29 & 0.9858 & 0.0103 & 33.55 & 0.9765 & 0.0260
        & 0.0027 & 37.28 & 0.9773 & 0.0177 \\
        \midrule
        Hotdog & Opaque & 67k
        & 38.92 & 0.9791 & 0.0589 & 29.28 & 0.9532 & 0.0786
        & 0.0047 & 34.30 & 0.9629 & 0.0808 \\
         & Soft & 67k
        & 39.66 & 0.9791 & 0.0602 & 28.75 & 0.9494 & 0.0823
        & 0.0058 & 34.16 & 0.9609 & 0.0829 \\
        \midrule
        Jugs & Opaque & 46k
        & 45.24 & 0.9930 & 0.0135 & 44.60 & 0.9862 & 0.0397
        & 0.0004 & 40.15 & 0.9870 & 0.0194 \\
         & Soft & 46k
        & 46.12 & 0.9943 & 0.0112 & 44.77 & 0.9875 & 0.0365
        & 0.0005 & 40.45 & 0.9879 & 0.0186 \\
        \midrule
        Average & Opaque & 65k
        & 40.78 & 0.9822 & 0.0308 & 35.10 & 0.9667 & 0.0476
        & 0.0022 & 36.99 & 0.9726 & 0.0419 \\
         & Soft & 65k
        & 41.92 & 0.9844 & 0.0269 & 35.30 & 0.9681 & 0.0449
        & 0.0025 & 37.12 & 0.9728 & 0.0391 \\
        \bottomrule
    \end{tabular*}
\end{table*}

\paragraph{Rendering time analysis.}
\Cref{tab:supp_soft_geometry_profile} reports per-frame CUDA-event timings
for Soft and Opaque from separate profiling runs under the same relighting settings.
The measurements exclude acceleration-structure construction and environment preprocessing.
The largest time reduction occurs in the fused tracer, which includes intersection and sorting,
transmittance accumulation, secondary albedo queries, and secondary PBR shading.
Primary rasterization and attribute evaluation also become faster,
while lighting-ray sampling and primary PBR shading have similar costs.
\textit{Other} includes vertex preparation, pixel and output packing,
color compositing, and metadata processing.

\begin{table*}[!t]
    \centering
    \caption{Relighting time per frame (ms) for Soft and Opaque on Synthetic4Relight.
    Rasterization includes primary material and normal evaluation.
    Tracing includes intersection, sorting, transmittance accumulation,
    secondary albedo queries, and secondary PBR shading.
    PBR denotes primary surface shading.
    These render-only CUDA-event timings are collected in separate profiling runs
    and exclude acceleration-structure construction and environment preprocessing.}
    \label{tab:supp_soft_geometry_profile}
    \small
    \setlength{\tabcolsep}{3pt}
    \begin{tabular*}{\textwidth}{@{\extracolsep{\fill}}ll*{7}{r}@{}}
        \toprule
        Scene & Variant & Rasterization & Sampling & Tracing & PBR & AA & Other & Total \\
        \midrule
        Air balloons & Soft & 47.77 & 8.80 & 769.89 & 7.25 & 0.00 & 2.50 & 836.21 \\
         & Opaque & 5.52 & 8.65 & 33.66 & 6.90 & 1.21 & 2.17 & 58.10 \\
        \midrule
        Chair & Soft & 35.13 & 6.49 & 755.50 & 5.42 & 0.00 & 2.09 & 804.64 \\
         & Opaque & 4.11 & 6.19 & 28.42 & 5.21 & 0.76 & 1.81 & 46.49 \\
        \midrule
        Hotdog & Soft & 52.38 & 10.09 & 1119.32 & 7.74 & 0.00 & 2.21 & 1191.73 \\
         & Opaque & 4.36 & 9.85 & 46.00 & 7.57 & 0.89 & 1.82 & 70.50 \\
        \midrule
        Jugs & Soft & 25.85 & 5.82 & 585.07 & 4.79 & 0.00 & 1.65 & 623.18 \\
         & Opaque & 4.46 & 5.59 & 26.35 & 4.84 & 0.84 & 1.65 & 43.73 \\
        \bottomrule
    \end{tabular*}
\end{table*}

\section{Analysis of hybrid visibility}
\label{sec:supp_hybrid_visibility}

During triangle annealing, abrupt visibility changes near shadow boundaries
may induce normal rotations to compensate for photometric errors.

We blend per-triangle visibility using the normalized weights
$w_m=\alpha_m T_m/\sum_{j=1}^{M}\alpha_j T_j$ from \cref{eq:triangle_compositing}.
The binary queries $\nu_m,\nu_{\mathrm{D}}\in\{0,1\}$ use the triangle and blended normals,
respectively, with one indicating environment visibility and zero indicating mesh occlusion.
The deferred and hybrid specular components for one color channel are
\begin{equation}
    \begin{aligned}
        c_{\mathrm{D}}
        &= \nu_{\mathrm{D}}\bar{S}^{\mathrm{dir}}
            +(1-\nu_{\mathrm{D}})\bar{S}^{\mathrm{ind}}, \\
        c_{\mathrm{H}}
        &= \bar{\nu}\bar{S}^{\mathrm{dir}}+(1-\bar{\nu})\bar{S}^{\mathrm{ind}}.
    \end{aligned}
    \label{eq:visibility_shading_comparison}
\end{equation}
Here $\bar{\nu}=\sum_m w_m\nu_m$ is the accumulated visibility,
and $\bar{S}^{\mathrm{dir}}$ and $\bar{S}^{\mathrm{ind}}$ are direct and indirect
specular responses evaluated with blended geometry and materials.
Both formulations share the same diffuse term.
Hybrid visibility lets both specular contributions and their weights adjust jointly,
reducing reliance on normal rotation.

\begin{figure}[!htb]
    \centering
    \includegraphics[width=\linewidth]{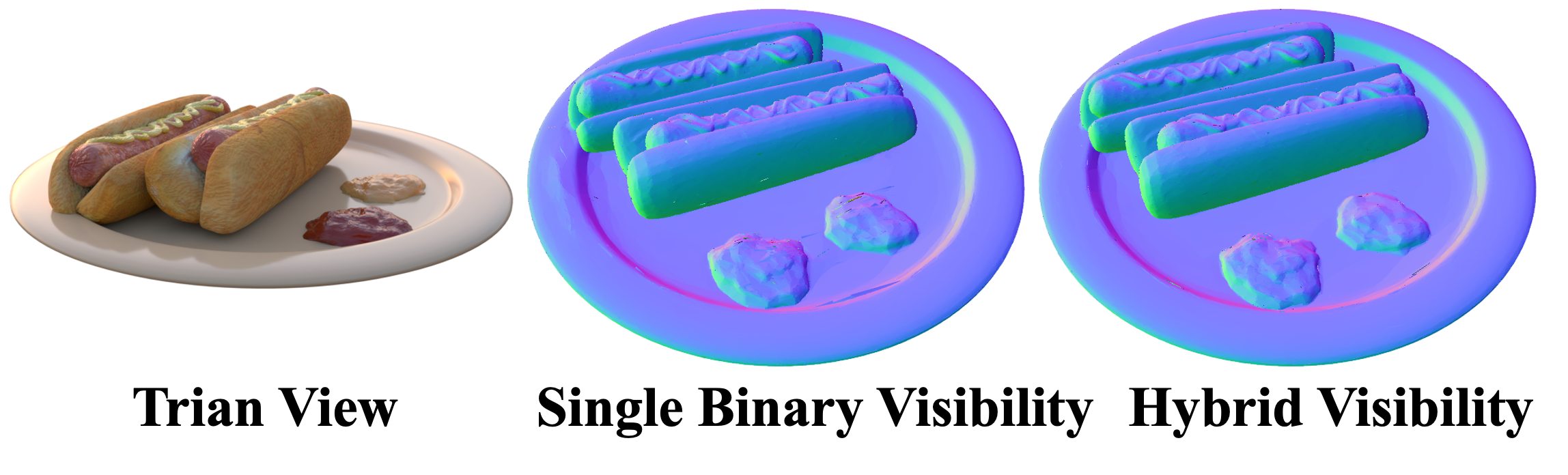}
    \caption{Hybrid visibility ablation on the Hotdog scene.
    Normal maps rendered after the first stage compare single and hybrid visibility.}
    \label{fig:ablation_hybrid_visibility}
\end{figure}

\Cref{tab:supp_hybrid_visibility} compares single and hybrid visibility on TensoIR.
The training view in \cref{fig:ablation_hybrid_visibility}
shows pronounced specular variations along the plate rim.
These variations can introduce artifacts in the rendered normal maps near the boundary,
as seen in the result with single visibility.
Hybrid visibility partially alleviates these errors in such regions,
but reconstruction of highly specular objects remains challenging during annealing.

\begin{table*}[!ht]
    \centering
    \caption{Hybrid visibility ablation on TensoIR, averaged equally over four scenes.}
    \label{tab:supp_hybrid_visibility}
    \small

    \begin{tabular*}{\textwidth}{@{\extracolsep{\fill}}l*{10}{c}@{}}
        \toprule
        & \multicolumn{3}{c}{Novel view synthesis} & \multicolumn{3}{c}{Albedo}
        & \multicolumn{3}{c}{Relighting} & Normal \\
        \cmidrule(lr){2-4}\cmidrule(lr){5-7}\cmidrule(lr){8-10}
        & PSNR$\uparrow$
        & SSIM$\uparrow$
        & LPIPS$\downarrow$
        & PSNR$\uparrow$
        & SSIM$\uparrow$
        & LPIPS$\downarrow$
        & PSNR$\uparrow$
        & SSIM$\uparrow$
        & LPIPS$\downarrow$
        & MAE ($^\circ$)$\downarrow$ \\
        \midrule
        Single visibility
        & 37.06 & 0.9702 & 0.0445 & 34.43 & 0.9456 & 0.0786
        & 32.74 & 0.9483 & 0.0623 & 3.659 \\
        \textbf{Hybrid visibility}
        & 37.10 & 0.9703 & 0.0444 & 34.43 & 0.9457 & 0.0784
        & 32.75 & 0.9484 & 0.0621 & 3.653 \\
        \bottomrule
    \end{tabular*}
\end{table*}

\section{Further analysis of triplane albedo}
\label{sec:supp_triplane_albedo}

\begin{table*}[!t]
    \centering
    \caption{Albedo representation analysis on Chair.
    Direct training uses 20k iterations. DiffSoup uses $R=5$.
    Fitting converts the trained triplane to vertex albedo or DiffSoup textures.
    Relighting averages two environment maps.}
    \label{tab:supp_triplane_analysis}
    \small

    \begin{tabular*}{\textwidth}{@{\extracolsep{\fill}}l*{10}{c}@{}}
        \toprule
        Method & \multicolumn{3}{c}{Albedo} & Roughness
        & \multicolumn{3}{c}{Novel view synthesis} & \multicolumn{3}{c}{Relighting} \\
        \cmidrule(lr){2-4}\cmidrule(lr){6-8}\cmidrule(lr){9-11}
        & PSNR$\uparrow$ & SSIM$\uparrow$ & LPIPS$\downarrow$ & MSE$\downarrow$
        & PSNR$\uparrow$ & SSIM$\uparrow$ & LPIPS$\downarrow$
        & PSNR$\uparrow$ & SSIM$\uparrow$ & LPIPS$\downarrow$ \\
        \midrule
        \multicolumn{11}{@{}l}{\textit{Direct training}} \\
        Vertex
        & 31.11 & 0.9422 & 0.0686 & 0.0032
        & 40.06 & 0.9665 & 0.0441 & 35.91 & 0.9646 & 0.0456 \\
        DiffSoup
        & 30.87 & 0.9524 & 0.0522 & 0.0039
        & 39.30 & 0.9730 & 0.0244 & 34.78 & 0.9668 & 0.0307 \\
        \textbf{Triplane}
        & 32.87 & 0.9718 & 0.0311 & 0.0029
        & 41.56 & 0.9825 & 0.0158 & 37.01 & 0.9760 & 0.0214 \\
        \midrule
        \multicolumn{11}{@{}l}{\textit{Triplane fitting}} \\
        Vertex
        & 32.51 & 0.9404 & 0.0700 & 0.0029
        & 39.82 & 0.9638 & 0.0482 & 36.63 & 0.9635 & 0.0481 \\
        DiffSoup
        & 32.87 & 0.9718 & 0.0311 & 0.0029
        & 41.56 & 0.9825 & 0.0159 & 36.98 & 0.9759 & 0.0220 \\
        \bottomrule
    \end{tabular*}
\end{table*}

\subsection{Comparison of albedo representations}
\label{sec:supp_albedo_representations}
We compare vertex albedo, per-triangle texture grids from
DiffSoup~\cite{tojo2026diffsoup} with $R=5$, and our shared triplane
on the Chair scene from Synthetic4Relight.
The direct variants use 20,000 training iterations.
We use the evaluation metrics in \cref{sec:experimental_settings}.

\begin{figure*}[!t]
    \centering
    \includegraphics[width=\textwidth]{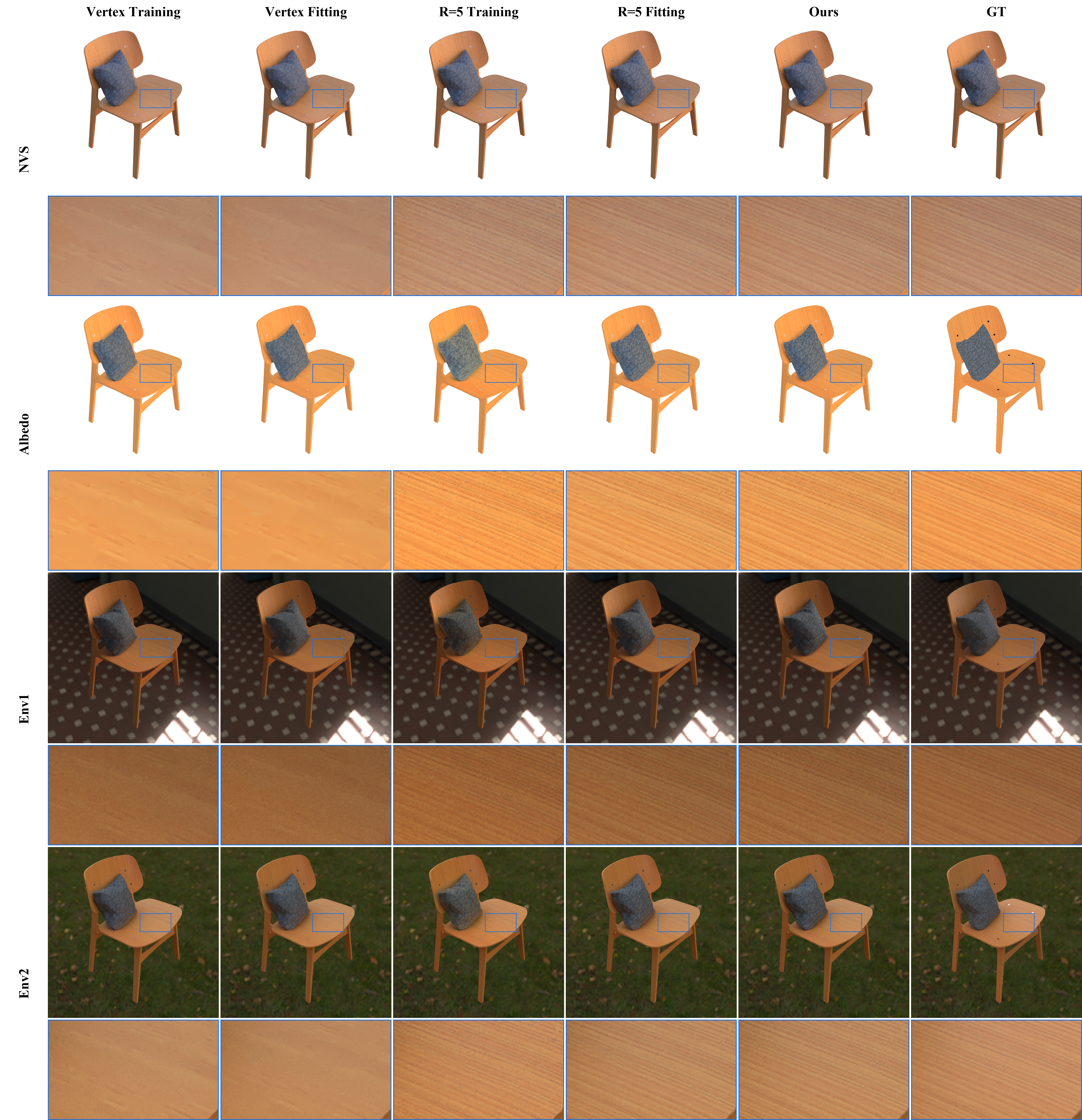}
    \caption{Albedo representation comparison on Chair.
    Columns show direct vertex training, vertex fitting from the triplane,
    direct DiffSoup training ($R=5$), DiffSoup fitting from the triplane,
    our triplane, and ground truth.
    Rows show novel view renderings, albedo, and relighting under two environment maps,
    with enlarged crops of the marked regions.}
    \label{fig:supp_triplane_analysis}
\end{figure*}

\Cref{tab:supp_triplane_analysis} and \cref{fig:supp_triplane_analysis}
compare the three directly trained representations and two conversions of triplane albedo.
The conversions fit vertex albedo and DiffSoup textures to the trained triplane
using bounded least squares, with all other parameters fixed.
They use neither ground truth nor test cameras and add no PBR optimization steps.
Direct triplane training achieves 32.87\,dB albedo PSNR,
compared with 31.11\,dB for vertex albedo and 30.87\,dB for DiffSoup.
Its higher albedo SSIM and lower LPIPS indicate better texture recovery,
with corresponding improvements in novel view and relighting quality.

\subsection{Fitting triplane albedo}
\label{sec:supp_triplane_fitting}
Fitted DiffSoup textures closely reproduce the triplane teacher,
indicating sufficient representational capacity in this scene.
Our triplane shares spatial features across nearby surface locations,
providing stronger spatial consistency during direct inverse rendering.
The results suggest that this spatial consistency improves albedo recovery.
Vertex fitting loses fine wood-grain details,
with lower albedo SSIM, higher LPIPS, and reduced novel view quality.

\section{Further analysis of anti-aliasing}
\label{sec:supp_anti_aliasing_analysis}

We further analyze the number of subpixel samples used for visibility estimation
and the search for native pixels that provide sample colors.

\begin{table*}[!t]
    \centering
    \caption{Anti-aliasing component ablations on Hotdog from TensoIR.
    Relighting uses the bridge environment.
    Sampling variants retain the complete color search;
    color-search variants use $4\times4$ samples per pixel.}
    \label{tab:supp_anti_aliasing_analysis}
    \small
    \setlength{\tabcolsep}{3pt}
    \begin{tabular*}{\textwidth}{@{\extracolsep{\fill}}l*{6}{c}@{}}
        \toprule
        Configuration & \multicolumn{3}{c}{Novel view synthesis}
        & \multicolumn{3}{c}{Relighting} \\
        \cmidrule(lr){2-4}\cmidrule(lr){5-7}
        & PSNR$\uparrow$ & SSIM$\uparrow$ & LPIPS$\downarrow$
        & PSNR$\uparrow$ & SSIM$\uparrow$ & LPIPS$\downarrow$ \\
        \midrule
        \multicolumn{7}{@{}l}{\textit{(a) Number of subpixel samples}} \\
        1 (w/o anti-aliasing)
        & 35.20 & 0.9702 & 0.0827 & 30.22 & 0.9268 & 0.1137 \\
        4 ($2\times2$)
        & 37.82 & 0.9754 & 0.0659 & 31.13 & 0.9355 & 0.1004 \\
        \textbf{16 ($4\times4$, Ours)}
        & 38.35 & 0.9767 & 0.0601 & 31.24 & 0.9372 & 0.0974 \\
        64 ($8\times8$)
        & 38.44 & 0.9769 & 0.0590 & 31.28 & 0.9375 & 0.0970 \\
        \midrule
        \multicolumn{7}{@{}l}{\textit{(b) Color selection strategy}} \\
        \textbf{Complete search (Ours)}
        & 38.35 & 0.9767 & 0.0601 & 31.24 & 0.9372 & 0.0974 \\
        w/o geometry/material scoring
        & 37.95 & 0.9757 & 0.0620 & 30.95 & 0.9350 & 0.1008 \\
        Nearest foreground pixel only
        & 37.39 & 0.9729 & 0.0706 & 30.60 & 0.9284 & 0.1069 \\
        \bottomrule
    \end{tabular*}
\end{table*}

\paragraph{Experimental settings.}
We use Hotdog from TensoIR and fix the model after 20,000 second-stage iterations.
Native PBR images are rendered at $800\times800$ with 768 lighting directions per pixel
and indirect illumination enabled, and are cached for all variants.
We evaluate novel view synthesis under the learned illumination
and relighting under the bridge environment.

\paragraph{Effect of subpixel sample number.}
We retain the complete color search and vary the uniform sampling grid
from $1\times1$ to $2\times2$, $4\times4$, and $8\times8$.
The single-sample configuration corresponds to rendering without anti-aliasing.
\Cref{tab:supp_anti_aliasing_analysis}(a) shows that increasing the grid
from $2\times2$ to $4\times4$ improves novel view and relighting PSNR
by 0.53 and 0.11\,dB, respectively.
Increasing it further to $8\times8$ gives smaller gains of 0.09 and 0.04\,dB,
with corresponding improvements in SSIM and LPIPS.
The quality gains therefore diminish beyond 16 samples in this scene.

\paragraph{Effect of color selection strategy.}
With a fixed $4\times4$ grid and coverage, we compare two simpler color selection strategies.
The first prefers pixels on the same triangle and otherwise uses the nearest foreground pixel.
The second always uses the nearest foreground pixel.
Our strategy improves all three metrics over both alternatives
(\cref{tab:supp_anti_aliasing_analysis}(b)).
Compared with the second strategy, novel view and relighting PSNR
increase by approximately 0.95 and 0.64\,dB, respectively.

\section{Limitations}
\label{sec:limitations}

TS+ annealing accelerates rendering by producing opaque triangles,
but reconstruction of highly specular objects remains challenging.
Despite partial vertex sharing, the topology remains fragmented,
leaving small triangles that are difficult to optimize.

\onecolumn
\clearpage
\section{Additional experimental results}
\label{sec:supp_additional_results}

\subsection{TensoIR}
\label{sec:supp_tensoir_results}
\begin{figure}[!ht]
    \centering
    \includegraphics[width=\textwidth]{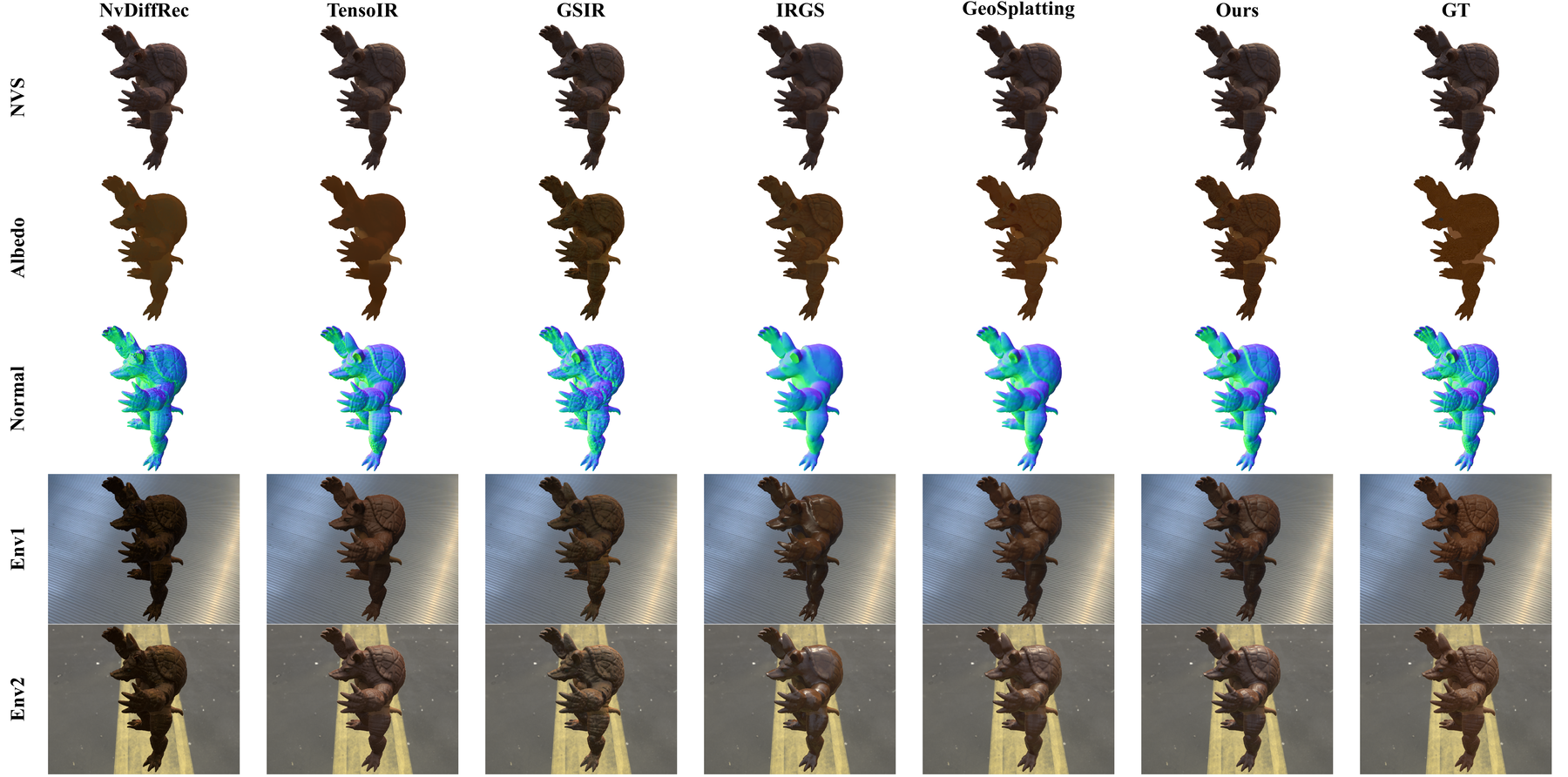}
    \caption{Comparison on the Armadillo scene from TensoIR.
    Rows show novel view renderings, albedo, normals, and relighting under two environments.
    The rightmost column shows the ground truth.}
    \label{fig:supp_tensorir_armadillo}
\end{figure}

\begin{figure}[!ht]
    \centering
    \includegraphics[width=\textwidth]{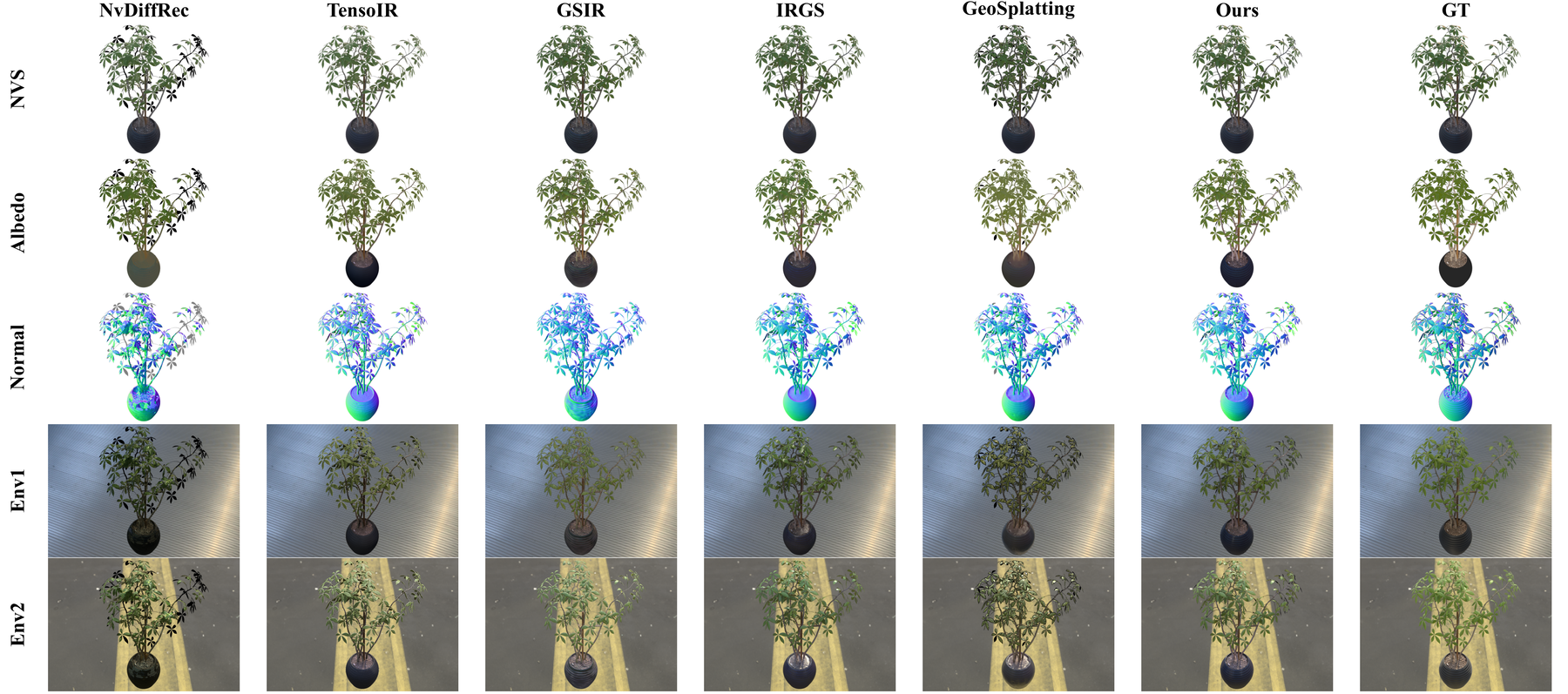}
    \caption{Comparison on the Ficus scene from TensoIR.
    Rows show novel view renderings, albedo, normals, and relighting under two environments.
    The rightmost column shows the ground truth.}
    \label{fig:supp_tensorir_ficus}
\end{figure}

\begin{figure}[!ht]
    \centering
    \includegraphics[width=\textwidth]{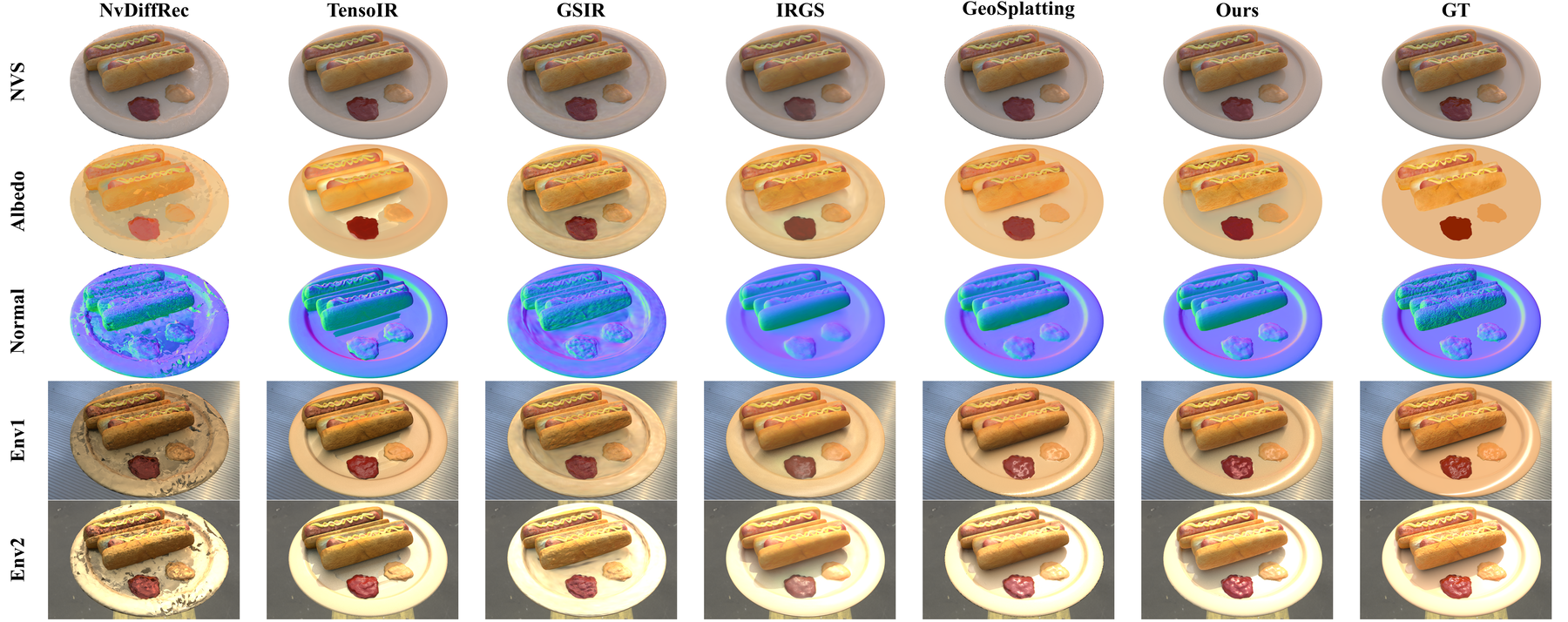}
    \caption{Comparison on the Hotdog scene from TensoIR.
    Rows show novel view renderings, albedo, normals, and relighting under two environments.
    The rightmost column shows the ground truth.}
    \label{fig:supp_tensorir_hotdog}
\end{figure}

\begin{figure}[!ht]
    \centering
    \includegraphics[width=\textwidth]{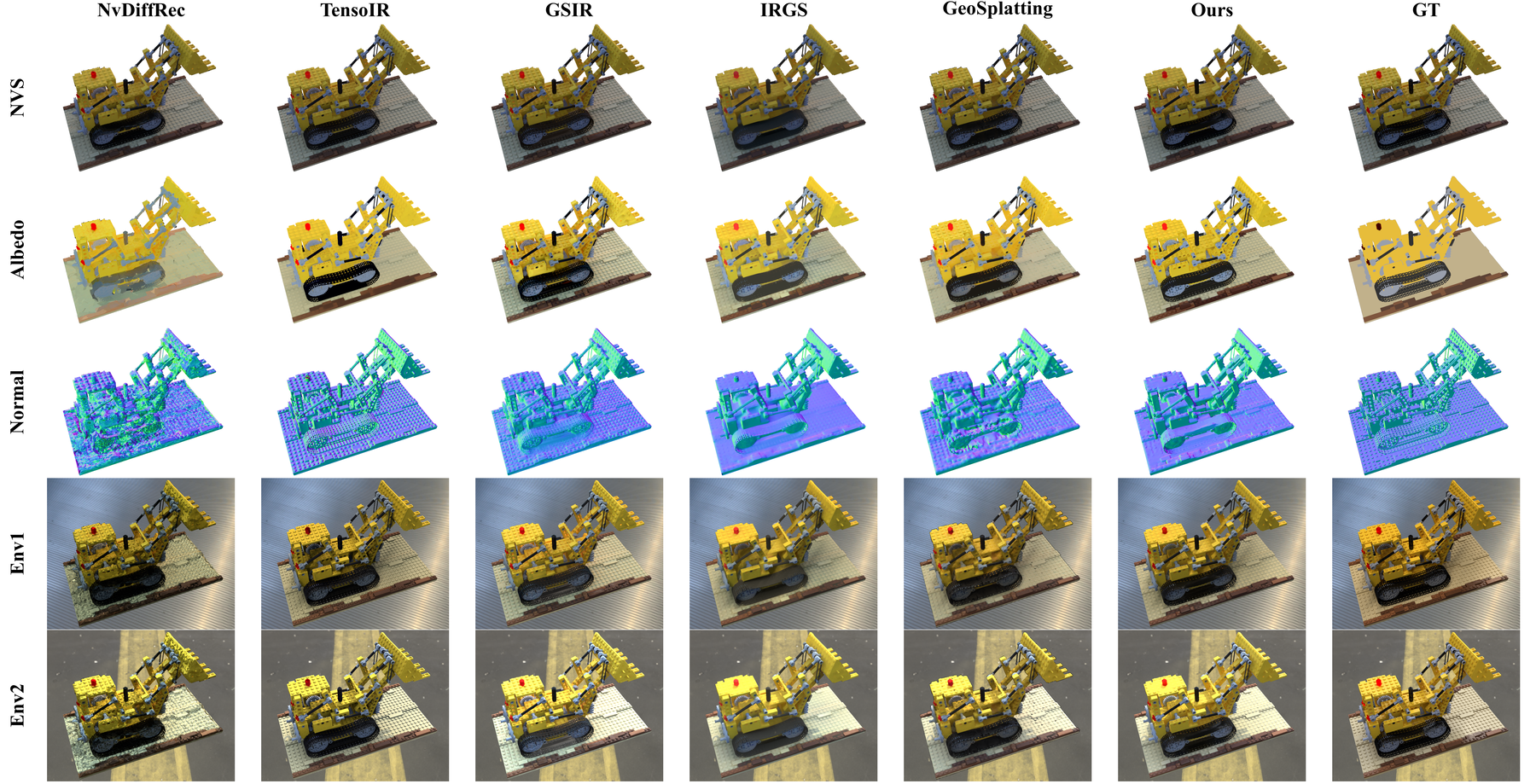}
    \caption{Comparison on the Lego scene from TensoIR.
    Rows show novel view renderings, albedo, normals, and relighting under two environments.
    The rightmost column shows the ground truth.}
    \label{fig:supp_tensorir_lego}
\end{figure}

\Cref{fig:supp_tensorir_armadillo,fig:supp_tensorir_ficus,fig:supp_tensorir_hotdog,fig:supp_tensorir_lego}
provide additional comparisons on the four TensoIR scenes.

\clearpage

\subsection{Synthetic4Relight}
\label{sec:supp_synthetic_results}
\begin{figure}[!ht]
    \centering
    \includegraphics[width=\textwidth]{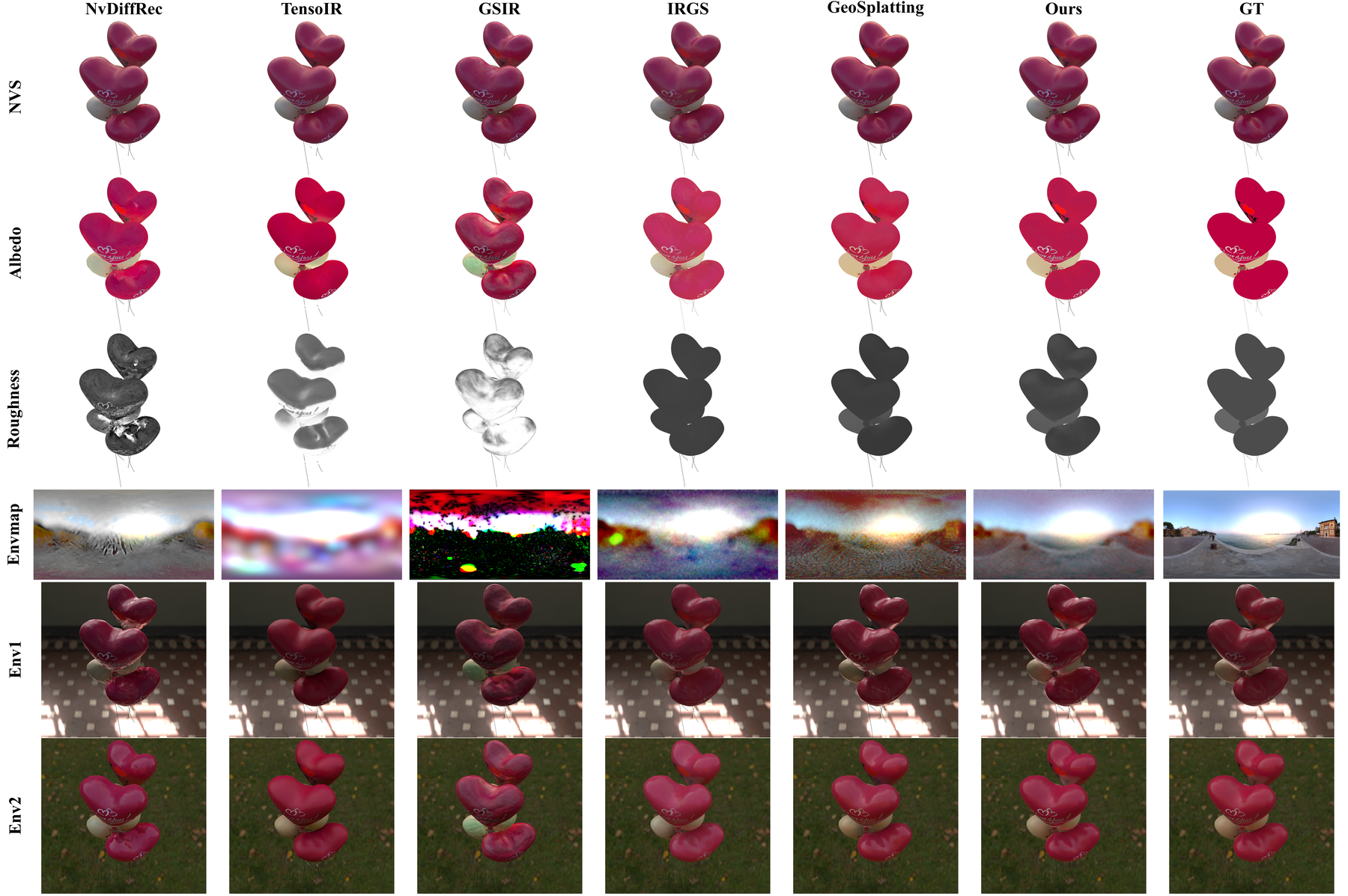}
    \caption{Comparison on the Air Balloons scene from Synthetic4Relight.
    Rows show novel view renderings, albedo, roughness, environment maps,
    and relighting under two environments.
    The rightmost column shows the ground truth.}
    \label{fig:supp_syn_air_baloons}
\end{figure}

\begin{figure}[!ht]
    \centering
    \includegraphics[width=\textwidth]{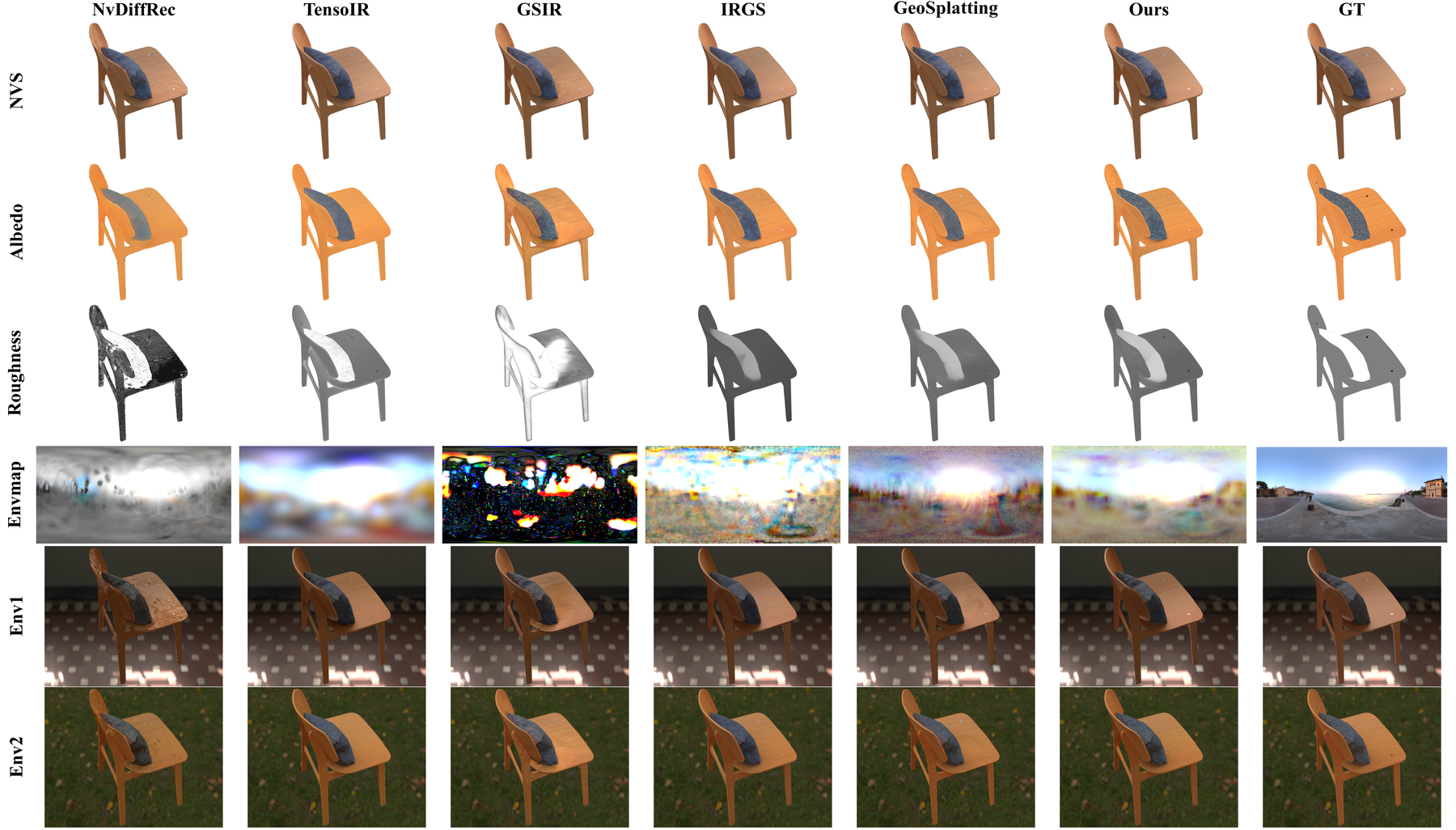}
    \caption{Comparison on the Chair scene from Synthetic4Relight.
    Rows show novel view renderings, albedo, roughness, environment maps,
    and relighting under two environments.
    The rightmost column shows the ground truth.}
    \label{fig:supp_syn_chair}
\end{figure}

\begin{figure}[!ht]
    \centering
    \includegraphics[width=\textwidth]{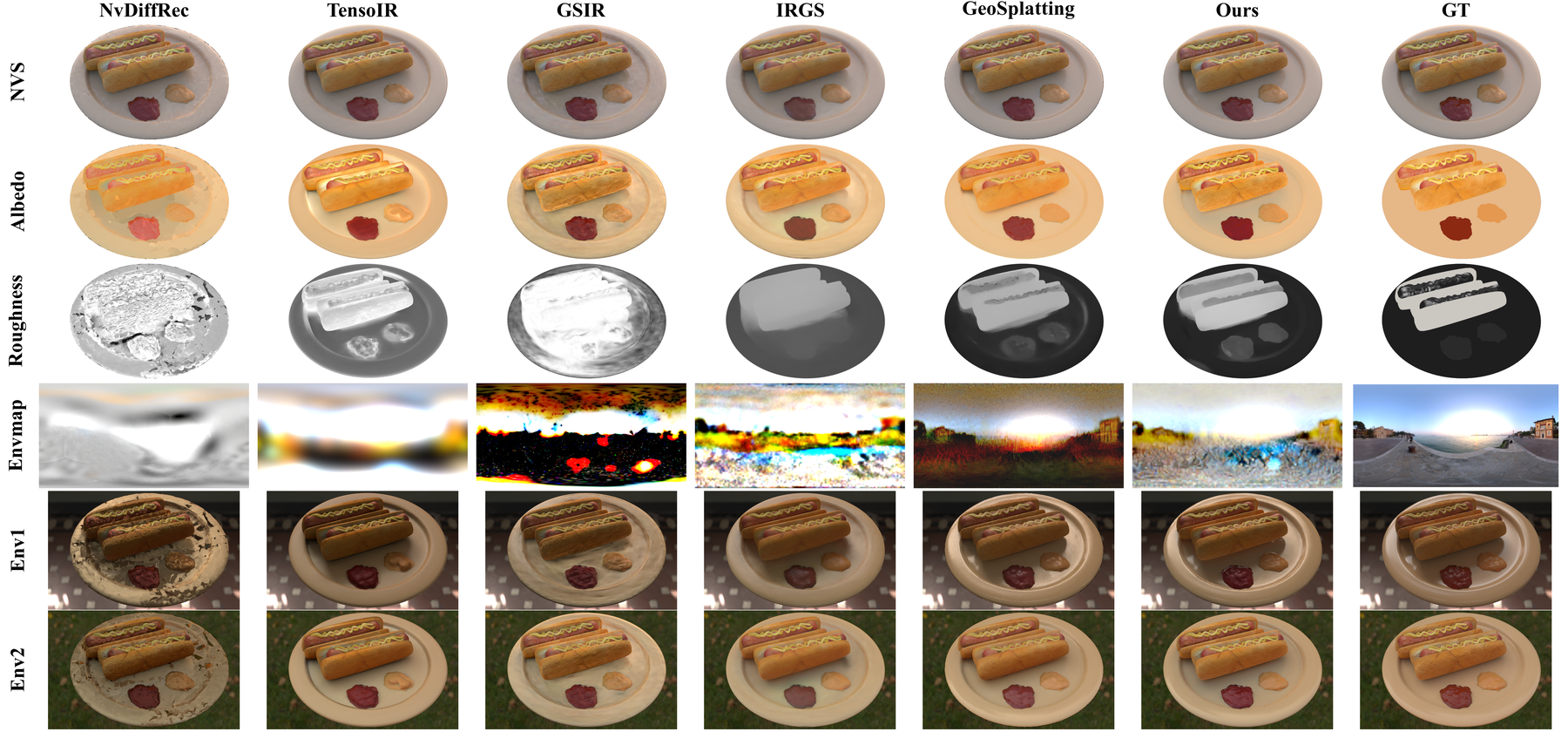}
    \caption{Comparison on the Hotdog scene from Synthetic4Relight.
    Rows show novel view renderings, albedo, roughness, environment maps,
    and relighting under two environments.
    The rightmost column shows the ground truth.}
    \label{fig:supp_syn_hotdog}
\end{figure}

\Cref{fig:supp_syn_air_baloons,fig:supp_syn_chair,fig:supp_syn_hotdog}
provide additional comparisons on Synthetic4Relight.

\clearpage

\end{document}